%% file: MAIN_arxiv.tex
\documentclass[11pt]{article}
\usepackage{longtable}
\usepackage{threeparttable}

\usepackage{graphicx} 
\usepackage{xcolor}
\usepackage{tablefootnote}   
\usepackage{enumitem}
\usepackage[singlelinecheck=false,labelfont=bf]{caption}
\usepackage[utf8]{inputenc}   % to be safe with special chars if using pdfLaTeX
\usepackage{amsmath}          % for nicer equations
\usepackage{float} 
\usepackage[numbers,sort&compress]{natbib}

\input{macros}

\date{}     
\title{Benchmarking Small Language Models and Small Reasoning Language Models on System Log Severity Classification}
\author{
    Yahya Masri \\
    George Mason University
    \and
    Emily Ma \\
    George Mason University
    \and
    Zifu Wang  \\
    Harvard University
    \and
    Joseph Rogers \\
    George Mason University
    \and
    Chaowei Yang\thanks{Email: {\tt cyang3@gmu.edu}} \\
    George Mason University
}

\begin{document}
\maketitle
\begin{abstract}
System logs are crucial for monitoring and diagnosing modern computing infrastructure, but their scale and complexity require reliable and efficient automated interpretation. Since severity levels are predefined metadata in system log messages, having a model merely classify them offers limited standalone practical value, revealing little about its underlying ability to interpret system logs. We argue that severity classification is more informative when treated as a benchmark for probing runtime log comprehension rather than as an end task. Using real-world journalctl data from Linux production servers, we evaluate nine small language models (SLMs) and small reasoning language models (SRLMs) under zero-shot, few-shot, and retrieval-augmented generation (RAG) prompting. The results reveal strong stratification. Qwen3-4B achieves the highest accuracy at 95.64\% with RAG, while Gemma3-1B improves from 20.25\% under few-shot prompting to 85.28\% with RAG. Notably, the tiny Qwen3-0.6B reaches 88.12\% accuracy despite weak performance without retrieval. In contrast, several SRLMs, including Qwen3-1.7B and DeepSeek-R1-Distill-Qwen-1.5B, degrade substantially when paired with RAG. Efficiency measurements further separate models: most Gemma and Llama variants complete inference in under 1.2 seconds per log, whereas Phi-4-Mini-Reasoning exceeds 228 seconds per log while achieving $<$10\% accuracy. These findings suggest that (1) architectural design, (2) training objectives, and (3) the ability to integrate retrieved context under strict output constraints jointly determine performance. By emphasizing small, deployable models, this benchmark aligns with real-time requirements of digital twin (DT) systems and shows that severity classification serves as a lens for evaluating model competence and real-time deployability, with implications for root cause analysis (RCA) and broader DT integration.
\end{abstract}

\section{Introduction}
System logs are vital components of modern computing infrastructure, capturing operational events, warnings, and performance information across distributed systems \cite{gholamian2021comprehensive, du2017deeplog, sekar2024eaudit, albert2025system}. They play a critical role in diagnosing faults, monitoring system health, and supporting automated responses in large-scale environments such as data centers and digital twins (DTs) \cite{viola2022combining, dietz2020integrating, zhang2017syslog}. However, as computing systems generate massive volumes of logs with complex, context-dependent language, manual interpretation has become infeasible \cite{xu2009detecting, jiang2009understanding, landauer2023deep, gupta2023learning, mahindru2021log}. The continuous flow of log entries produced by servers far exceeds human review capacity, leading to delayed fault detection and increased mean time to resolve \cite{bansal2020decaf, sun2025accurate}.

Recent advances in combining large language models (LLMs) and retrieval-augmented generation (RAG) have demonstrated that external context can significantly improve model reliability on tasks that require domain awareness and factual grounding. Rather than relying solely on parameters learned during training, RAG allows models to incorporate relevant evidence at inference time by querying a knowledge database. This approach has shown strong benefits in knowledge-intensive applications such as question answering, structured information extraction, and technical reasoning, where precision and traceability are essential \cite{wang2025optimizing, masri2025comparative, setty2024improving}. In system operation settings, retrieval enables models to surface historical signals, recurring patterns, and contextual system metadata, creating the foundation for downstream diagnostic reasoning. As DT architectures increasingly integrate intelligence and autonomy, retrieval becomes a key mechanism for linking runtime observations to prior system states, enabling more interpretable and context-aware decision pipelines.

Events and messages originating from the kernel, applications, and users of a system are recorded in system logs. Thus, these logs form an extensive record of processes executed within a network \cite{yang2021semi}. This provides system administrators with a crucial resource for monitoring performance, detecting security threats, and conducting root cause analysis (RCA) \cite{chen2022bert, landauer2020system, yang2021semi}.

The Syslog protocol was created as a framework to allow machines to transmit these event notifications \cite{gerhards2009rfc}. Each Syslog message contains a PRI component that consists of a ``$<$'', a number, and a ``$>$''. The number is known as the severity value, which is a combination of the log's Facility and Severity values. The Facility value is an integer ranging from 0 to 23, which describes the context of the log. The Severity value is an integer from 0 to 7 that quantifies the risk level of each log. Multiplying the Facility value by 8 and adding the Severity yields the Priority, which administrators use to identify and resolve system errors \cite{lonvick2001rfc3164}. This lack of strict standardization limits the use of severity labels as a canonical ground truth. However, it also makes them a realistic and challenging probe for evaluating whether language models (LMs) can align log content with operational intent under ambiguity \cite{el2020systematic}.

Within DT-oriented monitoring pipelines, such log interpretation must be both accurate and latency-efficient, motivating the evaluation of compact, deployable LMs under strict output and runtime constraints. 

The study focuses on evaluating small language models (SLMs) and small reasoning language models (SRLMs) using log severity classification as a controlled probe of their ability to ground real-world system log semantics under constrained outputs, based on logs collected from journalctl within the computing infrastructure.

\section{Related work}\label{sec:related}

\subsection{Log Classification with Manual Methods}\label{sec:manual}

In the past, developers created sets of rules for processes such as anomaly detection, leading to manually implemented systems. However, with the rapid development of computing infrastructure in both complexity and scale, these methods have become time-consuming and error-prone \cite{meng2021logclass}. Modern computing systems generate logs rapidly; for example, \citet{le2022log} estimate a rate of 30-50 gigabytes (about 120—200 million lines) per hour for Alibaba’s email production cloud computing system. The massive volume of system logs produced is impossible to manually traverse, complicating efforts to uncover patterns and nuanced faults that contribute to system failures \cite{zhu2019tools}. 

\citet{he2021survey} observe another notable limitation: developers may suffer from insufficient technical expertise, such as an understanding of runtime behaviors, the functions of different log levels, or best logging practices. Often, large-scale systems will have hundreds of contributors, with each individual having deep knowledge of one sub-component of the overall system. Due to this specialization, developers may have gaps in their understanding of overall system behaviors and relationships \cite{he2016experience}. Both conditions lead to struggles with assigning the proper log level. As a result, developers often must retrace their work and revise the levels assigned to previous system logs, further decreasing the efficiency \cite{li2021deeplv}.

\subsection{Log Classification with Traditional ML Methods}\label{sec:ml}

Due to the challenges of manual log classification, autonomous analysis methods have been explored through the creation of traditional machine learning (ML) methods \cite{dixit2022utilizing, henriques2020combining, zhang2024end}. Commonly used ML algorithms include random forest (RF) and support vector machine (SVM) \cite{ali2025comprehensive}.

Recently, \citet{qi2022adanomaly} utilized RF as a baseline, observing strong performance on Hadoop Distributed File System (HDFS), OpenStack, and BlueGene/L (BGL) datasets. The RF model often outperforms other classical ML methods, such as principal component analysis (PCA) and invariant mining (IM). Similarly,  \citet{li2022swisslog} evaluate SVM as a baseline and observe strong performance across the BGL and HDFS datasets, with the model achieving an F1-score of 0.96 and 0.92, respectively.

\citet{chen2021experience} observe that while ML methods represented a significant development for log analysis, key limitations prevent their widespread adaptation for practical use. Particularly, these methods often require the set of log events to be known beforehand, leading to difficulty in accounting for previously unseen events or sequences. As systems are upgraded and altered over time, many methods struggle to address new log events or changes in log semantics \cite{liu2023logbd}.

Another notable limitation of traditional ML models is their limited ability to model meaningful temporal dependencies between log events, including cause-related event sequences and abnormal timing patterns \cite{duan2024logedl}. As a result, critical information such as abnormally long delays or unusual event sequences is often not taken into consideration.

\subsection{Log Classification with DL Methods}\label{sec:dl}

More recently, deep learning has been explored to improve autonomous system log classification and analysis \cite{alzu2025cyberattack, ramachandran2023automated, yuan2020ada}. \citet{guo2024logformer} propose Logformer, a Transformer-based framework that is pre-trained on source domain log data and applied to the target domain via shared parameters, enhancing cross-domain performance. Logformer achieved higher F1 scores than baselines on the HDFS, BGL, and Thunderbird datasets.

\citet{li2024graph} propose Log2Graphs, a graph neural network model that detects graph-level anomalies for log anomaly detection. Compared to traditional baseline methods for ML and DL, Logs2Graphs performed better or similar across five datasets (HDFS, BGL, High-Performance Computing (HPC), Zookeeper, and Proxifier) as indicated by the area under the receiver operating characteristic curve (ROC AUC) and the area under the precision-recall curve (PRC AUC) measurements.

\citet{li2021deeplv} propose DeepLV, an automated DL approach for log level suggestion, outperforming ordinal regression (OR) and one-hot encoding recurrent neural network (RNN) baselines with respect to accuracy and AUC. Notably, this approach applies ordinal encoding to log levels in order to accurately represent sequential relationships between logs. The approach demonstrates best performance when combining both log message features and syntactic context, revealing the log's location.

\citet{wang2021multi} present OC4Seq, a multi-scale one-class RNN for anomaly detection. OC4Seq was tested on the HDFS, BGL, and RUBiS system log datasets. It was compared against five anomaly detection baselines: PCA, IM, One-Class SVM (OC-SVM), DeepLog, and DeepSVDD. OC4Seq demonstrates strong performance, achieving the highest F1 score across all three datasets with an F1 score of 0.976, 0.747, and 0.985 for HDFS, BGL, and RUBiS, respectively.

However, \citet{yu2024deep} caution that complex DL methods may not always be more accurate or time-efficient than traditional ML methods, especially when considering the greater computational costs of DL models. DL methods still suffer from some of the same issues that ML methods contend with, such as poor generalization across different datasets due to differences in log syntax and limitations on the quantity of labeled logs available for training. As such, creating a model that can be applied across different software systems poses a significant challenge \cite{bhanage2021infrastructure}. Currently, both ML and DL models often require manual intervention to optimize performance, especially for preprocessing and parameter tuning \cite{zhang2023system}. 

\subsection{Log Classification using LLMs}\label{sec:llms}

Several recent works have focused on leveraging LMs specifically for log classification, demonstrating significant improvements over traditional methods. \citet{yang2025logllama} propose LogLLaMA, a LLaMA2-based framework that first learns normal log patterns by predicting next log key sequences, then refines anomaly detection with a reinforcement learning objective using Top-$k$ rewards. Across BGL, HDFS, and Thunderbird, LogLLaMA consistently surpasses traditional and prior deep learning baselines, showing stronger robustness to unstable or unstructured logs and better generalization to unseen patterns.

\citet{zhang2025novel} present a GPT-based log anomaly detection framework that fine-tunes GPT 2 on structured log sequences derived from Drain parsed event IDs, then optimizes training with Focal Loss to handle extreme class imbalance. This optimization improves precision, recall, and F1 over vanilla GPT-2 while matching GPT-3.5 performance with efficient local deployment.

Additionally, \citet{huang2025logrules} introduce LogRules, a lightweight log analysis framework designed to reduce hallucinations in LLMs by teaching them explicit reasoning rules from logs. The system operates in three stages: induction, where a large model like GPT-4o-mini generates and validates log-related rules; alignment, where smaller models ($\sim$ 8B parameters) are fine-tuned via contrastive preference optimization to better apply those rules; and deduction, where rule-enhanced prompts guide inference on unseen logs. Across Loghub datasets, LogRules outperforms case-based LLM prompting and traditional methods in both log parsing and log-based anomaly detection, improving F1 by over 30\% on benchmarks like BGL and Spirit. This demonstrates that explicit rule induction and alignment substantially enhance reasoning accuracy and generalization for smaller LLMs in system log analysis.

Furthermore, recent studies have systematically evaluated the effectiveness of LMs in automating log-level classification tasks within software systems. \citet{heng2025benchmarking} benchmarked twelve open-source LLMs for log-level classification of Java source-code logging statements, comparing Fill-Mask and text-generation models across zero-shot, few-shot, and fine-tuned settings. Their results showed modest effectiveness, with baseline zero-shot and few-shot prompting often yielding accuracies below 30\%. Even with fine-tuning the best models (e.g., GraphCodeBERT, CodeLlama) plateaued around 60–70\% accuracy, highlighting ongoing challenges in deploying LLMs for static code-embedded log statement classification.

Building on the limitations identified in prior work, \citet{ouatiti2025omnillp} introduce OmniLLP, a retrieval-augmented framework that replaces random in-context examples with examples drawn from clusters that reflect real project structure. Source files are grouped by semantic similarity using code embeddings and by developer ownership using version control history, and in-context examples are retrieved from these coherent neighborhoods. Evaluated on four large open-source Java systems, this cluster-informed retrieval improves AUC by up to 8\% over random and documentation-based baselines, and the combined semantic-plus-ownership setting achieves AUC scores between 0.88 and 0.96, showing the effectiveness of incorporating software engineering context into LLM-based log level prediction.

Current research on LMs for log analysis remains fragmented and does not yet provide a comprehensive benchmark for real-world operational system logs. Many studies rely on Java logging statements or code-contextual logs rather than production system logs \cite{heng2025benchmarking}, while others focus on cybersecurity or intrusion-detection logs that differ substantially from general Linux operational logs \cite{karlsen2024benchmarking}. Additional work centers on telecom, network-device, or vendor-specific logs \cite{ihalage2025convolutional}, limiting generalizability to common system environments. Several influential benchmarks use structured datasets such as BGL, HDFS, or Thunderbird \cite{liu2024interpretable, ocansey2025logtinyllm}, which do not capture the temporal structure, irregularity, and noise of real-world journalctl logs. Other studies examine log-generation tasks tied to source code instead of operational log interpretation \cite{kim2025automated}. 

Prior work comparing prompting strategies and RAG in log and text analysis remains limited in scope and misaligned with the requirements of system-log severity classification. Existing studies in intrusion detection and security contexts compare zero-shot, few-shot, RAG, and fine-tuning approaches, but they rely on proprietary intrusion datasets rather than operational Linux logs \cite{bui2024systematic,yang2025large}. Additional work on microservices and industrial pipelines investigates prompt engineering techniques such as minimal instruction, Chain-of-Thought (CoT), and few-shot prompting, but focuses on log summarization, not classification \cite{ahmed2025optimizing}. Even within log-specific literature, prompting-strategy studies emphasize anomaly detection, log parsing, and summarization rather than classification tasks, and use heavily preprocessed and template-parsed HPC or distributed-system logs as BGL, HDFS, and Thunderbird instead of journalctl logs \cite{liu2024logprompt,cui2024logeval}. While recent work has explored RAG in log analysis, these studies primarily examine RAG's vulnerability to noisy, outdated, or heterogeneous log data such as noise-sensitive knowledge sources \cite{duan2025eagerlog}, retrieval errors and hallucination failure modes \cite{sharma2025retrieval}, and instability arising from multi-source or semi-structured logs \cite{wu2025multirag}, but they do not analyze how retrieval affects model performance in the context of system log severity classification using SLMs and SRLMs.

Although prior work reports inference latency and throughput for LLM-based log analysis, including per-log inference times for classification models \cite{karlsen2024benchmarking}, parsing time comparisons between small and large LLMs \cite{ma2025adaptivelog}, and efficiency evaluations within broader benchmark suites \cite{cui2024logeval}, these studies focus on anomaly detection, log parsing, summarization, or specialized domains such as telecom logs, and none evaluate inference latency or real-time performance for severity classification on Linux system logs, nor do they measure latency specifically for SLMs or SRLMs. Moreover, none of these efficiency-oriented studies incorporate RAG or analyze the additional inference latency overhead introduced by retrieval when combined with reasoning models, independent of predictive accuracy.

\subsection{Research Objectives}\label{sec:researchObj..}

Despite rapid progress in applying LMs to log analysis, existing research remains fragmented across log types, tasks, and modeling paradigms. Prior work has emphasized Java logging statements, security-oriented logs, telecom logs, or heavily preprocessed anomaly-detection datasets, leaving real-world Linux system logs (e.g., journalctl) comparatively understudied. Likewise, prompting-strategy research has largely centered on anomaly detection, summarization, or intrusion detection, and has not evaluated retrieval-induced degradation or the behavior of small or reasoning models on operational logs. Efficiency-oriented studies measure inference latency for parsing or anomaly detection, but no prior work assesses deployable SLM/SRLM performance or end-to-end latency for severity classification on journalctl data.

To address these gaps, this work makes the following contributions:

\begin{enumerate}
    \item We introduce the first benchmark evaluating SLMs and SRLMs on real-world Linux system logs (journalctl) for severity classification, in contrast to prior work focusing on code logging statements, anomaly datasets (e.g., HDFS, BGL), or security-specific logs.
    \item We compare zero-shot, few-shot, and retrieval-augmented prompting specifically for system log severity classification on journalctl data, and document retrieval-induced degradation effects for several SRLMs.
    \item We analyze both accuracy and inference latency for SLMs and SRLMs, providing deployability-oriented evaluation for real-time DT and monitoring pipelines.
\end{enumerate}

By systematically evaluating SLMs and SRLMs for severity classification on temporal-based operational Linux system logs under zero-shot, few-shot, and RAG approaches, this study establishes a foundation for future work on lightweight, real-time, and retrieval-aware log analysis, all important aspects that support the development of intelligent, automated infrastructure monitoring systems such as DTs.

\section{Data}\label{sec:data}

To develop a reliable benchmark for log severity classification, we collected over 7.3 million system log entries from six servers across the computing infrastructure—four internal nodes and two external public-facing machines. The logs were extracted from each system’s journal and converted into a structured JSON format, with each entry tagged by its originating IP address to maintain source traceability. To ensure consistent interpretation by LMs, every record was standardized into a key–value dictionary representation. Table 1 illustrates an example of a single log entry and its corresponding JSON structure. This structured design aligns with recent findings showing that explicit text formatting improves LLM comprehension and reliability \cite{braun2025hidden, he2024does}.

\begin{table}[H]
  \centering
  \caption{Example JSON-style representation of a system log entry.}
  \label{tab:json-example}
  \begin{tabular}{p{0.25\linewidth}p{0.65\linewidth}}
    \hline
    \textbf{Attribute} & \textbf{Example} \\
    \hline
    \texttt{id} & \texttt{57010} \\
    \texttt{hostname} & \texttt{ray-worker4} \\
    \texttt{ip} & \texttt{10.192.20.11} \\
    \texttt{comm} & \texttt{cat} \\
    \texttt{cmdline} & \texttt{/bin/cat} \\
    \texttt{exe} & \texttt{/usr/bin/cat} \\
    \texttt{message} &
      \texttt{-rw-r--r-- 2 root root 0 Oct 26 2021 usr/lib/kbd/keymaps/legacy/mac/all/mac-de\_CH.map.gz} \\
    \texttt{selinux\_context} &
      \texttt{unconfined\_u:unconfined\_r:rpm\_script\_t:s0-s0:c0.c1023} \\
    \texttt{systemd\_unit} & \texttt{session-1.scope} \\
    \texttt{systemd\_slice} & \texttt{user-0.slice} \\
    \texttt{realtime\_datetime} & \texttt{2025-02-14 11:38:10.852717} \\
    \texttt{priority} & \texttt{7.0} \\
    \hline
  \end{tabular}
\end{table}

The dataset spans from June 2024 to July 2025, capturing a broad range of operational conditions, from low-level kernel and hardware diagnostics to service activity and security notifications. Each entry retains critical metadata, including timestamp, process ID, command, message content, and priority level, which are essential features for accurate severity classification.

For manageable experimentation, we sampled 50,000 log entries at random from the full corpus of 7.3 million system logs, maintaining statistical representativeness while ensuring computational feasibility. The original corpus exhibited a pronounced class imbalance: informational messages (priority 6) dominated, while high-severity categories such as alert (1), critical (2), and error (3) occurred infrequently. To preserve the fidelity of rare but operationally significant events, all logs from priority levels 1–4 (alert–warning) were fully retained, yielding 16,635 entries. The remaining 33,365 samples were evenly drawn from lower-severity levels 5–7 (notice, informational, debug) to reduce bias toward frequent categories. After deduplication, the dataset contained 46,774 unique logs, reflecting the removal of 3,226 duplicates.

Following preprocessing, the dataset was divided into training and evaluation subsets using an 80/20 split with a fixed random seed to ensure reproducibility. The training subset contained 37,419 logs while the evaluation subset contained 9,355 logs. Stratified sampling preserved proportional representation of all severity levels across both subsets. To prevent data leakage, the target label corresponding to log priority was removed from the evaluation set prior to inference, producing a label-free version used for model testing.

Table 2 summarizes the final distribution of priority levels according to the standardized syslog scale, where level 0 denotes emergency (system unusable), levels 1–3 correspond to alert, critical, and error conditions, levels 4–5 indicate warning and notice messages, level 6 represents informational events, and level 7 corresponds to debug output. No emergency-level entries were observed in the collected logs—a reflection of the infrastructure’s stable operational state but also a limitation for evaluating model sensitivity to extreme conditions. 

\begin{table}[H]
\centering
\caption{Final distribution of 46,774 log entries across standardized syslog severity levels. Each level (0--7) represents increasing system stability from emergency to debug, illustrating a semi-balanced dataset that preserves coverage across both rare high-severity and common low-severity events observable in production systems.}
\label{tab:severity-distribution}
\begin{tabular}{l l r r}
\hline
\textbf{Severity Level} & \textbf{Description} & \textbf{Count} & \textbf{Percentage} \\
\hline
0 & Emergency & 0 & 0\% \\
1 & Alert     & 2,315 & 4.6\% \\
2 & Critical  & 5,412 & 10.8\% \\
3 & Error     & 7,025 & 14.1\% \\
4 & Warning   & 9,678 & 19.3\% \\
5 & Notice    & 8,744 & 17.5\% \\
6 & Info      & 10,286 & 20.6\% \\
7 & Debug     & 6,314 & 12.1\% \\
\hline
\end{tabular}
\end{table}

This semi-balanced, empirically representative benchmark provides a realistic foundation for evaluating SLMs and SRLMs under operational conditions, positioning log severity classification as a practical means to gauge model comprehension and its relevance to downstream tasks such as DT system integration.

\section{Methodologies}\label{sec:methodologies}

\begin{figure}[H]
    \centering
    \includegraphics[width=0.95\linewidth]{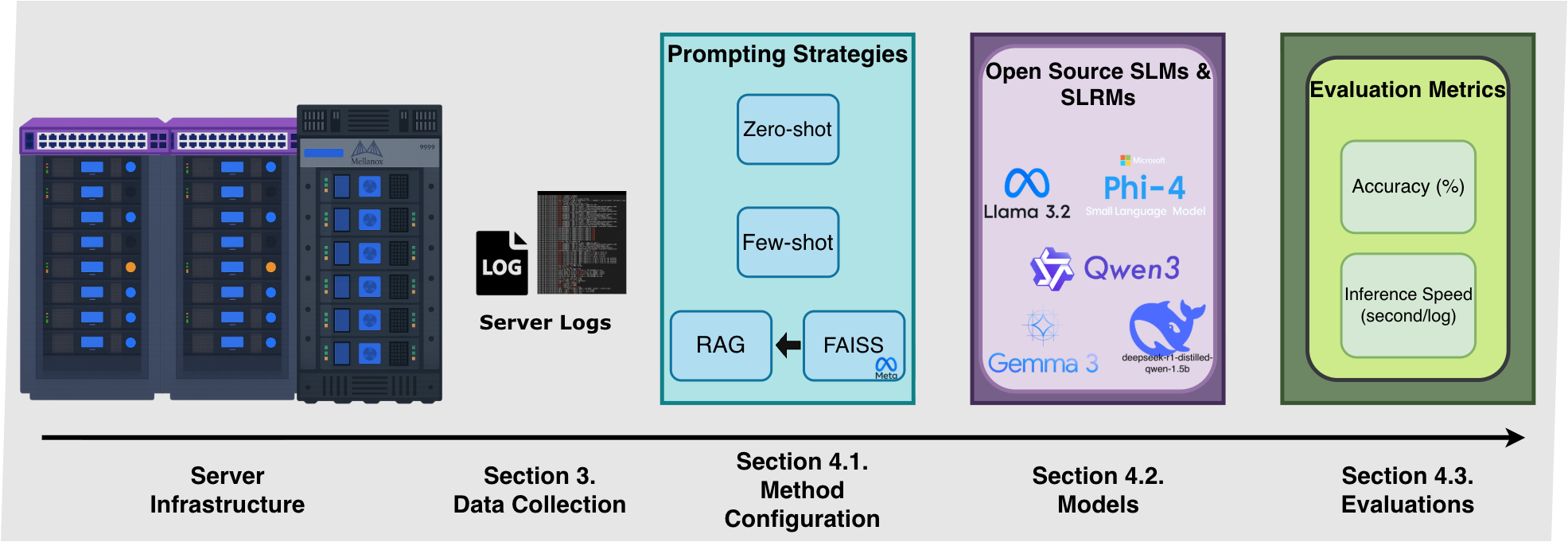}
    \caption{Workflow for evaluating log-severity classification using SLMs and SRLMs. System logs are collected from real-world servers, processed through prompting strategies (zero-shot, few-shot, and RAG with FAISS), evaluated across various open-source models, and assessed by accuracy and inference speed.}
    \label{fig:workflow}
\end{figure}

Figure 1 outlines the workflow for evaluating the performance of using different prompting methods and open-source small language models (SLMs) and small reasoning language models (SRLMs) to classify system log severity. After collecting over 7.3 million log entries from six servers within the computing infrastructure, each log was standardized into a structured JSON format and sampled to produce a semi-balanced benchmark for evaluation. Each entry was processed using three prompting methods: zero-shot prompting \cite{radford2019language, kojima2022large}, few-shot prompting \cite{brown2020language}, and retrieval-augmented generation (RAG) \cite{lewis2020retrieval}. Zero-shot and few-shot prompts were applied to infer the correct severity level from the underlying log content. For RAG implementation, Facebook AI Similarity Search (FAISS), an open-source library developed by Meta for efficient similarity search and clustering of dense vectors, was utilized to construct a vector database from the training set \cite{douze2025faiss}. During evaluation, each test log queried this database using the same prompt to retrieve contextually similar entries for severity inference. Each method was tested across multiple open-source models with consistent hyperparameter settings to ensure comparability. Final evaluation considered overall accuracy and average inference time per log entry to assess both model comprehension and runtime efficiency. More details are provided in the sub-sections.

\subsection{Method Configuration}\label{sec:mconfig}
\subsubsection{Zero-shot}\label{sec:zs}

The zero-shot configuration evaluated each model's ability to classify logs without any examples or prior context exposure. The prompt framed the model as a Linux System Log Specialist with expertise in system administration, log analysis, and troubleshooting critical system-level issues through comprehensive log examination as well as familiarity with journalctl syntax and Syslog standards. Following \citet{kong2024better}, this role-based framing was used to enhance zero-shot reasoning by providing the model with an explicit professional context that encourages domain-consistent interpretation and decision-making.

To enforce output consistency, a strict rule was applied: models were to produce only a single digit from 0 to 7 with no explanation, punctuation, or whitespace. This design choice standardized responses for efficient evaluation and eliminated ambiguity from model verbosity.

\subsubsection{Few-shot}\label{sec:fs}

The few-shot configuration expanded upon the baseline setup by including example logs with their corresponding Syslog severity level directly in the prompt to guide the model’s reasoning process. The prompt retained the same professional framing as in the zero-shot configuration, positioning the model as a Linux System Log Specialist with expertise in system administration, log analysis, and familiarity with journalctl syntax and Syslog standards. However, rather than relying solely on domain instructions, the model was provided five example log entries paired with their severity levels.

The five examples were sampled directly from the training subset to prevent data leakage and ensure that the evaluation set remained unseen. These examples were intentionally chosen to capture a range of operational conditions from routine informational messages to critical disk errors, allowing the models to observe the underlying linguistic and structural cues that distinguish high- and low-severity logs. Following \citet{dong2024survey}, the inclusion of in-context exemplars was designed to activate the model’s internal task priors, improving alignment between log semantics and severity classification through analogical reasoning. 

As mentioned prior, models were instructed to produce only a single integer from 0 to 7, without any accompanying explanation or punctuation. This consistency maintained direct comparability with the zero-shot configuration while isolating the performance gains derived from contextual exemplars.

\subsubsection{RAG}\label{sec:rag}

The retrieval-augmented configuration extended the zero-shot prompt by pairing it with a vector-based memory of the training subset, which contained 37,419 logs containing their Syslog severity level. This design aimed to enhance interpretability and classification precision through dynamic grounding in prior data rather than the static exemplars used in the few-shot configuration.

To construct the retrieval corpus, a FAISS index was built over the training set $D_{\text{train}} = \{(x_i, y_i)\}_{i=1}^{N}$ where $x_i$ denotes a structured log entry and $y_i \in \{0, 1, \ldots, 7\}$ denotes its Syslog severity label. Each indexed document was generated by converting a row from the training subset into a key-value text representation, where each log field (e.g., hostname, message, pid) was arranged into a Python dictionary-like string concatenated with its corresponding Syslog severity level, producing a semantic embedding $ \mathbf{e}_i = f_{\theta}(x_i) \in \mathbb{R}^{768} $, where $f_{\theta}$  denotes the Nomic Embed text encoder (\texttt{nomic-embed-text-v1.5}), which produces 768-dimensional embeddings and is held fixed across all experiments to ensure comparability \cite{nussbaum2024nomic}.

At inference time, each unseen $x_q$ was embedded into the same vector space and queries against the index using an L2 similarity metric. The retriever then selected the top-$k$ neighbors: 
\[
\mathcal{N}_{k}(x_q)
= \argmin\nolimits_{x_i \in D_{\text{train}}}
\lVert \mathbf{e}_q - \mathbf{e}_i \rVert_2
\]

For all experiments, $k$=5 was used as the default retrieval depth. The retrieved examples were automatically formatted as in-context snippets appended below the zero-shot instruction prompt. This configuration thus combined the interpretability of structured retrieval with the generalized capacity of zero-shot prompting. By conditioning on $\mathcal{N}_{k}(x_q)$, the model effectively approximated a similarity-weighted local decision function: 
\[
\hat{y}_q = g_{\phi}\!\left( x_q,\, \mathcal{N}_{k}(x_q) \right)
\]
where $g_{\phi}$ denotes the LM's classification mapping under RAG conditioning. This approach enabled more informed classification by using prior knowledge from semantically aligned logs.

\begin{figure}[H]
    \centering
    \includegraphics[width=0.95\linewidth]{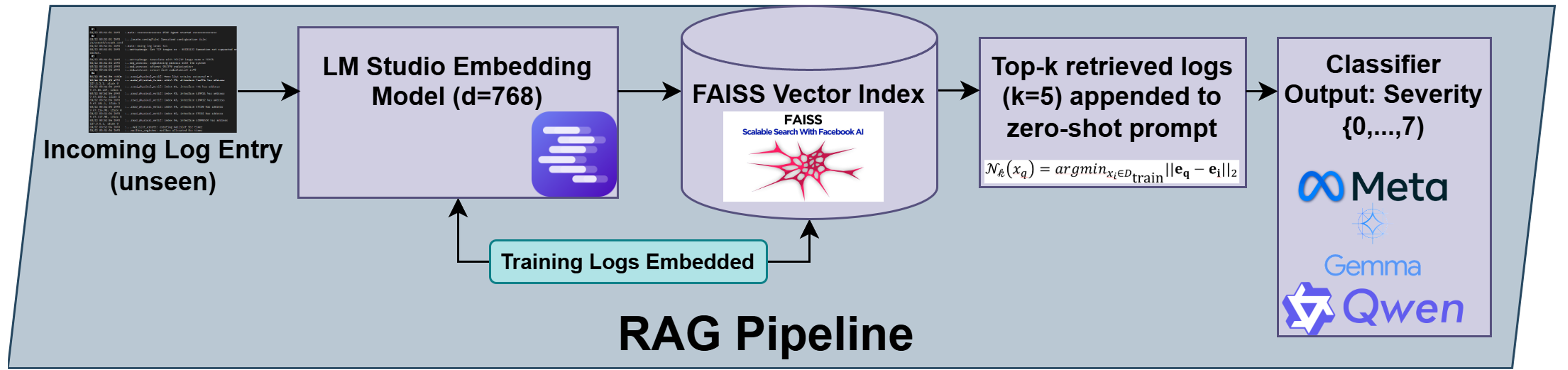}
    \caption{Retrieval-Augmented Generation (RAG) Pipeline for Syslog severity classification where incoming logs are embedded using a local LM Studio encoder (768-dimensional embeddings) and indexed with FAISS. At inference, each query log is embedded, and the top-$k$ nearest training vectors ($k$=5) are retrieved using L2 similarity and appended to a zero-shot prompt. The LLM then predicts the Syslog severity label conditioned on the retrieved examples.}
    \label{fig:rag}
\end{figure}

\subsubsection{Hardware and Environment}\label{sec:h&w}

All experiments were executed on a high-performance local workstation configured for large-scale inference. The system was equipped with an Intel Xeon W3-2423 processor, 128 GB RAM, and dual NVIDIA RTX A6000 GPUs (each providing 48 GB of VRAM). This setup offered sufficient inference speed, acceleration, and capacity for dense vector retrieval.

Model evaluation was hosted through LM Studio, which provided an offline OpenAI-compatible API endpoint for serving the S(R)LMs used in this study \cite{lmstudio2025}. This hardware-software arrangement ensured reproducibility across experiments involving different prompting configurations.

\subsection{Models}\label{sec:models}

SLMs are compact neural models designed for efficient inference, lower computational demand, and reduced energy use, making them suitable for constrained deployment environments \cite{wang2025survey}. Recent benchmarking shows that SLMs can achieve strong task performance while providing measurable gains in runtime efficiency and power consumption across a range of settings, including multi-domain evaluation suites and diverse hardware platforms \cite{pham2025slm}. A NVIDIA study defines SLMs functionally as models that can be served on common personal devices with latency sufficiently low for practical use, noting that models below roughly ten billion parameters are currently capable of this in practice \cite{belcak2025small}. Following this definition, this study focuses on the 0.6B–4B parameter range to ensure deterministic execution, stable latency, and full local control over inference conditions, aligning with current research emphasis on reproducibility and resource-bounded evaluation \cite{belcak2025small}. This scale also enables integration within DT environments, remaining practical for real-time pipelines that demand fast, continuous, and locally-reliable inference.

In parallel, SRLMs extend the SLM paradigm by inheriting reasoning through distillation from larger models, enabling multi-step reasoning while maintaining smaller computational footprints \cite{wang2025short}. These models leverage CoT supervision, reward-guided refinement, and test-time scaling strategies to improve reasoning without increasing parameter count, reflecting a broader trend toward efficiency-oriented cognitive modeling \cite{wang2025short}. Architectural developments such as grouped-query attention (GQA), linear-time sequence models like Mamba, and lightweight normalization layers further enhance SLM efficiency by reducing memory requirements and inference latency while preserving expressivity \cite{wang2025short}. By evaluating both SLMs and SRLMs, we assess whether distilled reasoning signals provide additive performance benefits in resource-constrained classification settings and whether compact architectures can deliver consistent, domain-relevant accuracy under tight latency and compute budgets.

To evaluate the generalization and reasoning capacity of LMs under different prompting strategies, a diverse set of small-scale models spanning two distinct categories was selected: SLMs and SRLMs. The SLM group comprises variants of the Llama3.2 and Gemma3 families, optimized primarily for efficient natural-language understanding and instruction following with minimal context. In contrast, the SRLM group includes variants of the Qwen3 family and other models like DeepSeek-R1-Distill-Qwen-1.5B and Phi-4-Mini-Reasoning, each incorporating CoT aimed at improving step-by-step inference and logical consistency.

The models evaluated in this study are summarized in Table 3, grouped by their classification as either SLMs or SRLMs.

\begin{table}[H]
\centering
\caption{Models evaluated in this study, categorized as either small language models (SLMs) or small reasoning language models (SRLMs).}
\label{tab:models}
\begin{tabular}{l c}
\hline
\textbf{Model} & \textbf{Category} \\
\hline
Llama3.2-3B & SLM \\
Llama3.2-1B & SLM \\
Gemma3-4B & SLM \\
Gemma3-1B & SLM \\
Qwen3-4B & SRLM \\
Qwen3-1.7B & SRLM \\
Qwen3-0.6B & SRLM \\
DeepSeek-R1-Distill-Qwen-1.5B & SRLM \\
Phi-4-Mini-Reasoning (3.8B) & SRLM \\
\hline
\end{tabular}
\end{table}

Beyond their architectural goals, these models also differ in attention-head configuration and context capacity, critical factors for latency, retrieval performance, and real-time operation. 

\begin{table}[H]
\centering
\caption{Technical specifications of the evaluated models, including the number of query (Q) and key--value (KV) attention heads and maximum context length supported.}
\label{tab:tech_specs}
\begin{tabular}{lccc}
\hline
\textbf{Model} & \textbf{Q Heads} & \textbf{KV Heads} & \textbf{Max Context Length} \\
\hline
Llama3.2-3B & 24 & 8 & 131{,}072 \\
Llama3.2-1B & 32 & 8 & 131{,}072 \\
Gemma3-4B & 8 & 4 & 131{,}072 \\
Gemma3-1B & 4 & 1 & 32{,}768 \\
Qwen3-4B & 32 & 8 & 32{,}768 \\
Qwen3-1.7B & 16 & 8 & 32{,}768 \\
Qwen3-0.6B & 16 & 8 & 32{,}768 \\
DeepSeek-R1-Distill-Qwen-1.5B & 12 & 2 & 131{,}072 \\
Phi-4-Mini-Reasoning (3.8B) & 24 & 8 & 131{,}072 \\
\hline
\end{tabular}
\end{table}

Together, these models offer a balanced scope suitable for efficiency and potential real-time DT integration.

\subsection{Evaluation}\label{sec:eval}

To assess and evaluate the ability of SLMs and SRLMs on interpreting system logs, we evaluate each model across the three prompting configurations mentioned in Section 4.1: zero-shot, few-shot, and RAG. Evaluation is conducted on the test subset of the data, comprising unique log entries that were not shown during instruction or retrieval phases to prevent information leakage. For each configuration, models receive raw log text and are required to output a single Syslog severity level, enabling direct comparison of predictive accuracy across architectures and scales.

Accuracy served as the primary evaluation metric, capturing the proportion of predictions that matched the reference Syslog severity labels assigned in the collected system logs. To quantify classification performance, we compute accuracy as the fraction of correctly predicted severity levels over all log entries in the test set. Formally,

\[
\text{Accuracy} = \frac{1}{N} \sum_{i=1}^{N} \mathbb{I}(\hat{y}_i = y_i),
\]

where \(N\) denotes the total number of log messages, \(y_i\) is the ground-truth Syslog severity level for the \(i\)-th entry, \(\hat{y}_i\) is the model’s predicted label, and \(\mathbb{I}(\cdot)\) is the indicator function that returns 1 when the classification matches the reference label and 0 otherwise.

In addition, per-log inference latency was measured and averaged across the test subset to capture realistic performance implications for continuous monitoring environments and DT pipelines. Together, these evaluation procedures quantify model performance across accuracy, latency, and retrieval-driven reasoning efficiency, providing insight into both linguistic competence and operational viability in real-time log analysis systems.

\section{Results}\label{sec:res}
\subsection{Zero-shot}\label{sec:reszs}

Table 5 summarizes zero-shot log severity classification results for all evaluated models, reporting both accuracy and average inference time per log entry. Notably, the Qwen3 family of models achieved the highest accuracy, with Qwen3-1.7B leading at 33.61\%, outperforming all other models and size variants in this setup. In contrast, smaller variants of Llama3 and Gemma demonstrated substantially lower accuracy (below 10\%), highlighting the challenge of zero-shot severity inference for smaller SLMs. Per-log inference times varied widely, with lightweight models such as Llama3.2-1B achieving 0.08 seconds per log, while reasoning models such as DeepSeek-R1-Distill-Qwen-1.5B and Qwen3-0.6B took several seconds per log. One observation from the zero-shot runs is that the Phi-4-Mini-Reasoning model, while achieving 9.20\% accuracy, frequently appended unnecessary explanations to its classifications (e.g., outputting reasoning or justifications after the answer), which decreased its overall accuracy as the zero-shot prompt explicitly mentioned to produce a single digit output corresponding to the Syslog severity level. Overall, these results establish a baseline performance and reveal practical trade-offs between speed and accuracy in the zero-shot setup.

\begin{table}[H]
\centering
\caption{Zero-shot performance of evaluated models.}
\label{tab:zeroshot}
\begin{tabular}{lcc}
\hline
\textbf{Model} & \textbf{Accuracy} & \textbf{Average (seconds/log)} \\
\hline
Llama3.2-3B & 8.11\% & 0.13 \\
Llama3.2-1B & 1.04\% & 0.08 \\
Gemma3-4B & 4.79\% & 0.18 \\
Gemma3-1B & 0.14\% & 0.12 \\
Qwen3-4B & 27.12\% & 3.53 \\
Qwen3-1.7B & \textbf{33.61\%} & 1.71 \\
Qwen3-0.6B & 27.45\% & 4.19 \\
DeepSeek-R1-Distill-Qwen-1.5B & 11.54\% & 4.07 \\
Phi-4-Mini-Reasoning (3.8B) & 9.20\% & 13.18 \\
\hline
\end{tabular}
\end{table}

\subsection{Few-shot}\label{sec:resfs}

Table 6 presents few-shot prompt results for all models, showing considerable improvements in accuracy over the zero-shot baseline for most architectures. The Qwen3-4B model achieved the best performance, reaching an accuracy of 56.01\%, substantially higher than its zero-shot result and indicating strong benefit from exposure to labeled examples. Gemma3-4B and Qwen3-1.7B also showed marked gains, achieving 41.06\% and 43.30\% accuracy respectively. In contrast, some smaller models, such as Llama3.2-1B and Phi-4-Mini-Reasoning, failed to improve and in fact produced 0\% accuracy under few-shot prompts, while also suffering significant increases in inference time per log—up to 39.33 and 23.43 seconds, respectively. This was due to these models not following directions and producing lengthy outputs with verbose reasoning, resulting in unclear answers. Compared to the zero-shot setting, these results demonstrate not only higher accuracy for larger and mid-scale models but also increased inference latency, especially for models less optimized for prompt-following or with limited capacity. This highlights that few-shot prompting can produce stronger classification ability, but model choice and operational constraints remain key considerations.

\begin{table}[H]
\centering
\caption{Few-shot performance of evaluated models.}
\label{tab:fewshot}
\begin{tabular}{lcc}
\hline
\textbf{Model} & \textbf{Accuracy} & \textbf{Average (seconds/log)} \\
\hline
Llama3.2-3B & 33.21\% & 0.19 \\
Llama3.2-1B & 0.00\% & 39.33 \\
Gemma3-4B & 41.06\% & 0.26 \\
Gemma3-1B & 20.25\% & 0.22 \\
Qwen3-4B & \textbf{56.01\%} & 8.35 \\
Qwen3-1.7B & 43.30\% & 3.03 \\
Qwen3-0.6B & 28.92\% & 18.74 \\
DeepSeek-R1-Distill-Qwen-1.5B & 17.63\% & 5.53 \\
Phi-4-Mini-Reasoning (3.8B) & 0.00\% & 23.43 \\
\hline
\end{tabular}
\end{table}

\subsection{RAG}\label{sec:resRAG}

Table 7 details the model performance using RAG, revealing dramatic improvements in accuracy for most architectures compared to zero- and few-shot prompting. In particular, Gemma3-1B and Gemma3-4B achieved accuracy rates of 85.28\% and 81.84\%, respectively, while Qwen3-4B reached a near-perfect 95.64\%, sharply outperforming its previous results. Smaller models benefited substantially from RAG as well, such as Llama3.2-1B, which improved to 37.37\% accuracy, and Qwen3-0.6B rose to 88.12\%. However, inference latency generally increased, especially for larger and reasoning models: Qwen3-4B required 7.14 seconds per log, and Phi-4-Mini-Reasoning again struggled with both accuracy (0\%) and extremely high latency (over 3 minutes per log). 

Interestingly, the RAG results reveal that not all models benefited from retrieval augmentation. In fact, Qwen3-1.7B, DeepSeek-R1-Distill-Qwen-1.5B, and Phi-4-Mini-Reasoning (3.8B) saw their performance decline sharply. Qwen3-1.7B dropped from 43.30\% (few-shot) down to 28.96\% (RAG), while DeepSeek-R1-Distill-Qwen-1.5B fell from 17.63\% to just 3.17\%, performing even worse than its zero-shot accuracy. Most strikingly, Phi-4-Mini-Reasoning failed to produce any correct classification with RAG (0\% accuracy), despite long inference times. These findings highlight that RAG can introduce additional challenges, especially for models that struggle to meaningfully incorporate retrieved context or are sensitive to more complex prompts. Such declines in accuracy highlight the importance of model-specific tuning and careful evaluation. 

These findings illustrate that while RAG can provide critical context and substantially improve classification performance for many models, its benefits are not universal. Careful evaluation of both specific model architectures and size variants is essential to determine the most suitable approach for a given domain-specific task.

\begin{table}[H]
\centering
\caption{Retrieval-Augmented Generation (RAG) performance of evaluated models.}
\label{tab:rag}
\begin{tabular}{lcc}
\hline
\textbf{Model} & \textbf{Accuracy} & \textbf{Average (seconds/log)} \\
\hline
Llama3.2-3B & 53.31\% & 0.90 \\
Llama3.2-1B & 37.37\% & 0.63 \\
Gemma3-4B & 81.84\% & 1.16 \\
Gemma3-1B & 85.28\% & 0.70 \\
Qwen3-4B & \textbf{95.64\%} & 7.14 \\
Qwen3-1.7B & 28.96\% & 0.89 \\
Qwen3-0.6B & 88.12\% & 2.75 \\
DeepSeek-R1-Distill-Qwen-1.5B & 3.17\% & 4.88 \\
Phi-4-Mini-Reasoning (3.8B) & 0.00\% & 228.07 \\
\hline
\end{tabular}
\end{table}

\section{Discussion}\label{sec:discussion}
\subsection{Impact of Top-k Retrieval on Model Stability}\label{sec:topK}

Although Qwen3-4B demonstrated strong performance gains from retrieval, the 1.7B variant exhibited the opposite behavior. To investigate whether this degradation stemmed from excessive retrieval context, an experiment was conducted in which the top-$k$ neighbors returned by FAISS were reduced from $k$=5 to $k$=3 to $k$=1. Figure 3 summarizes the effect of reducing retrieval depth on accuracy and latency. 

\begin{figure}[H]
    \centering
    \includegraphics[width=0.9\linewidth]{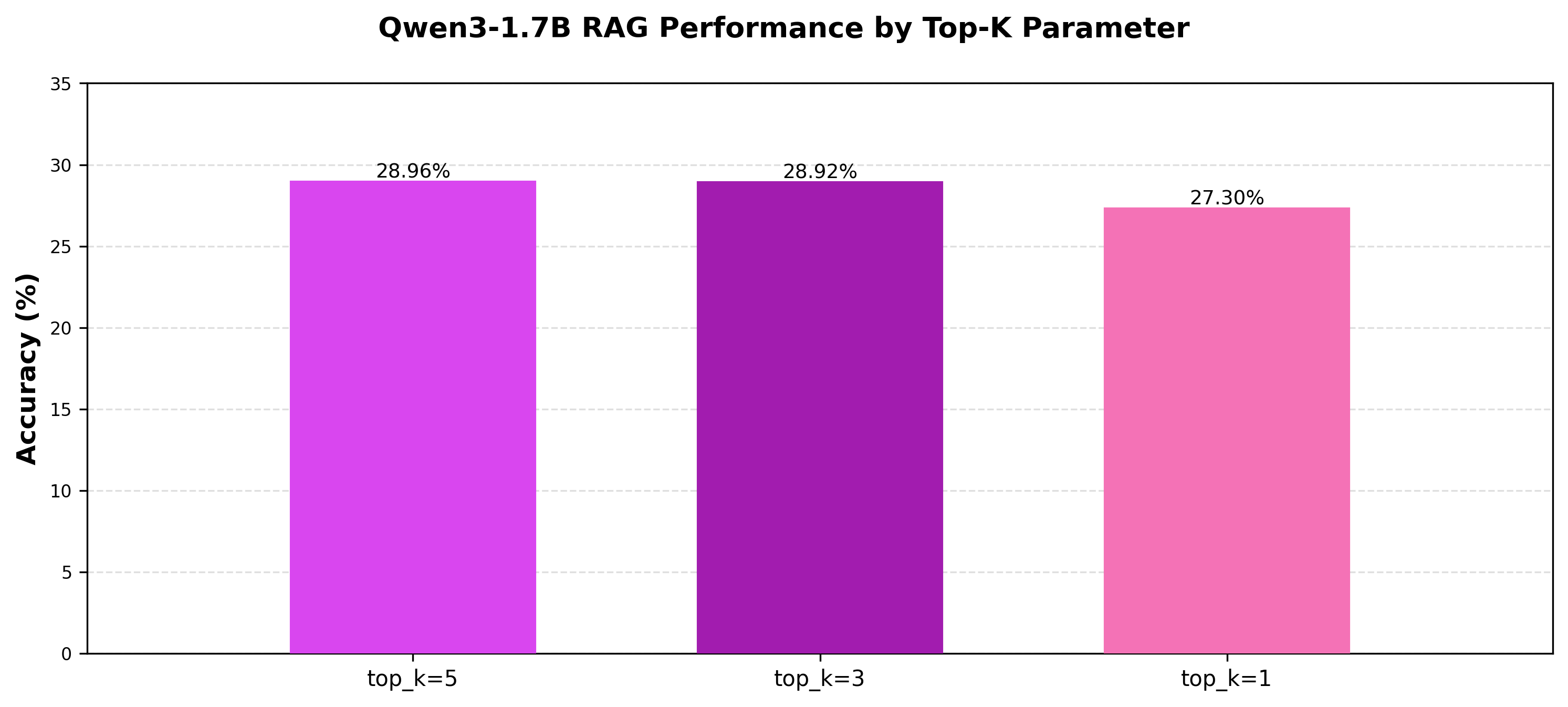}
    \caption{Performance of Qwen3-1.7B across retrieval depths. Reducing $k$ from 5 to 1 decreases latency but does not improve accuracy, indicating limited retrieval integration at this model scale.}
    \label{fig:topk}
\end{figure}

Results reveal that performance did not recover with smaller context windows when reducing the $k$ neighbors. Instead, accuracy declined steadily from 28.96\% at $k$=5 to 26.47\% at $k$=1, despite inference speed decreasing proportionally. 

This behavior suggests that the model’s challenge with retrieval is not driven by excessive context volume, but instead reflects a limitation in effectively incorporating retrieved information into its decision process. While retrieval substantially benefits certain compact models in our evaluation framework, these results indicate that gains do not scale uniformly across model sizes. Instead, successful retrieval integration appears to depend on a model’s capacity to absorb and utilize external evidence alongside internal representations. This effect is observed here within a domain-specific system log classification setting, and highlights retrieval-conditioning behavior as an important consideration when working with models near the lower-capacity regime.

\subsection{Overall Observations}\label{sec:obs}

Across all models and prompting strategies, we observe clear stratification in both accuracy and efficiency. RAG yields the greatest improvements, though the extent of these gains varies considerably by model family and scale. Figure 4 reports the model accuracy across the three different prompting configurations, while Figure 5 summarizes the per-log inference speed.

\begin{figure}[H]
    \centering
    \includegraphics[width=0.95\linewidth]{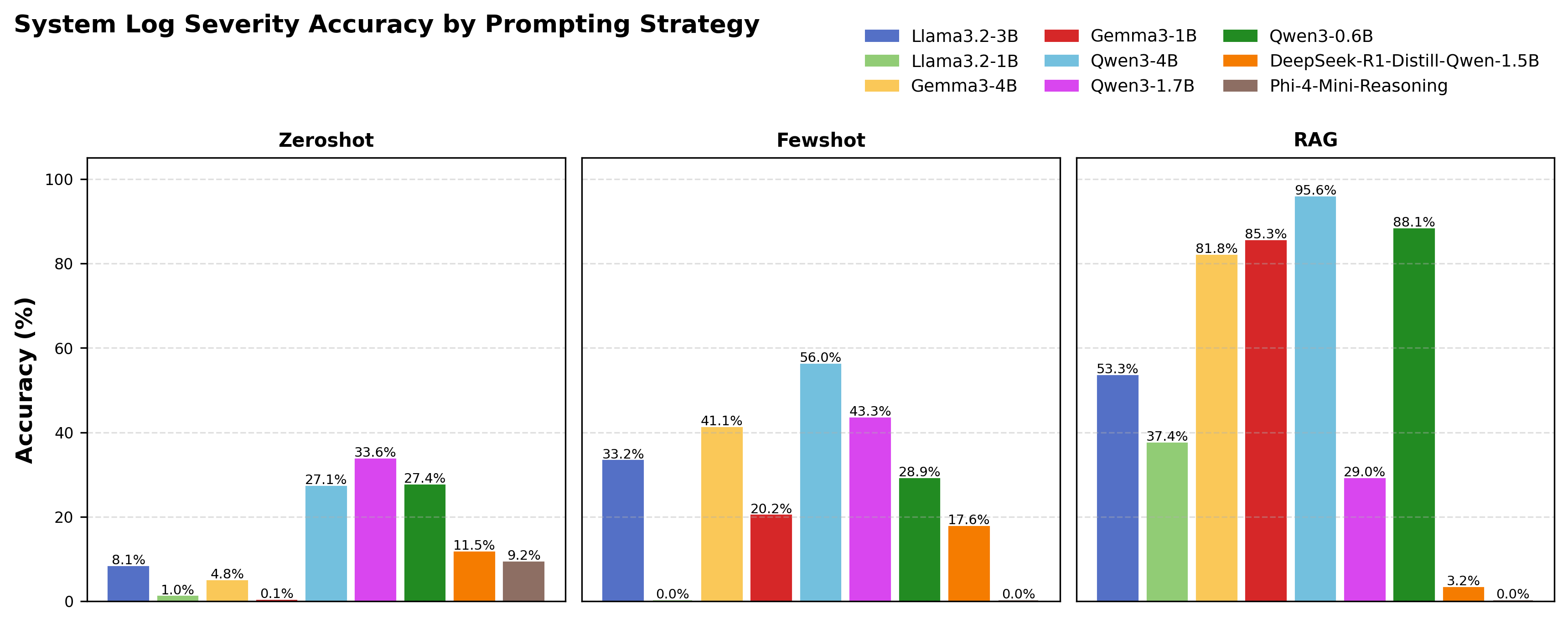}
    \caption{Performance of evaluated models under zero-shot, few-shot, and RAG prompting. Retrieval yields the strongest overall gains, with notable variation across architectures.}
    \label{fig:acc}
\end{figure}

\begin{figure}[H]
    \centering
    \includegraphics[width=0.95\linewidth]{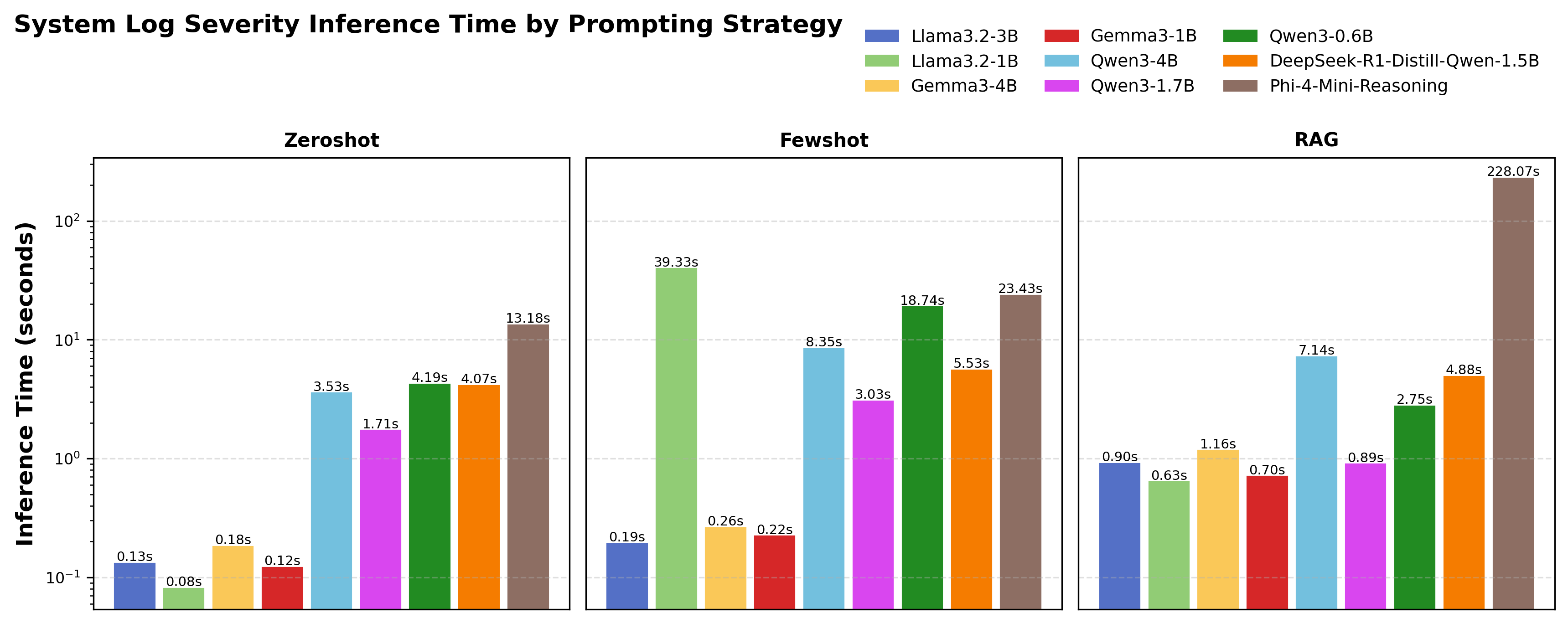}
    \caption{Average per-sample inference time. Several models approach real-time latency, though performance varies significantly by scale and prompting mode.}
    \label{fig:time}
\end{figure}

According to the results, Qwen3-4B is the top performer, reaching 95.64\% accuracy with RAG compared to 56.01\% in few-shot and 27.12\% in zero-shot. The Gemma family also shows substantial gains, with Gemma3-1B improving from 20.25\% (few-shot) to 85.28\% (RAG) and Gemma3-4B from 41.06\% to 81.84\%. Llama3.2 models benefit more modestly: Llama3.2-3B increases from 8.11\% (zero-shot) to 53.31\% (RAG), while Llama3.2-1B remains limited, peaking at 37.37\% under RAG and collapsing to 0.00\% in few-shot. Interestingly, Qwen3-0.6B emerges as an outlier, performing weakly in zero-shot (27.45\%) and few-shot (28.92\%) but jumping to 88.12\% with RAG, indicating that retrieval can elevate tiny models. In contrast, Qwen3-1.7B performs worse with RAG (28.96\%) than few-shot (43.30\%), and DeepSeek-R1-Distill-Qwen-1.5B degrades sharply from 17.63\% (few-shot) to 3.17\% (RAG). Phi-4-Mini-Reasoning fails across all settings, reaching a peak of 9.20\% in zero-shot and reaching 0.00\% in both few-shot and RAG. These anomalies suggest the possibility of structural incompatibilities between retrieval integration and some smaller or distilled architectures.

Processing time reveals different dynamics. Most models complete zero-shot and RAG inference in under a second per log, with Gemma variants particularly consistent below 1.2 seconds. Llama models are also efficient, though Llama3.2-1B produces an extreme outlier at 39.33 seconds in few-shot, caused by difficulty adhering to prompt instructions when example inputs and outputs were provided, which led to unnecessarily long generations. Qwen models incur greater latency: Qwen3-4B ranges from 3.53 to 8.35 seconds, Qwen3-1.7B from 0.89 to 3.03 seconds, and Qwen3-0.6B from 2.75 to 18.74 seconds. DeepSeek-R1-Distill-Qwen-1.5B falls in the middle range at 4–5.5 seconds. Phi-4-Mini-Reasoning is highly inefficient, requiring more than 228 seconds per log under RAG due to excessive $<$think$>$ token usage and verbose outputs, making it impractical despite its already poor accuracy.

Although several of the evaluated models achieved near–real-time performance, with 11 out of the 27 total evaluations completing inference in under one second, newer-generation GPUs with greater throughput and memory bandwidth would further reduce latency and bring the system closer to true real-time operation. These results are consistent with prior findings indicating that GPU acceleration plays a critical role, where rapid inference directly supports continuous synchronization with live system data. As prior studies show, inference time decreases significantly with higher GPU throughput and memory bandwidth \cite{chitty2024llm, wang2023adopting}.

\subsection{Architectural Factors Behind Performance Differences}\label{architecturedifferences}

The strong performance of the Qwen3 family, especially Qwen3-4B and the tiny Qwen3-0.6B, appears consistent with aspects of their underlying architecture and post-training pipeline. Qwen3 models are dense decoder-only Transformers with GQA and QK-Norm, trained on a very large, STEM and code-heavy corpus and then refined via a four-stage post-training process that jointly optimizes "thinking" (CoT) and "non-thinking" modes \cite{yang2025qwen3}. In our setting, Qwen3-4B combines a relatively large number of query heads (32 Q / 8 KV) with 32K context and strong reasoning/post-training, giving it both the capacity to project the mixed structure of logs (paths, numeric fields, timestamps, errors) into many attention subspaces and the stability to keep retrieved neighbors salient over long prompts. Qwen3-0.6B shares the same 16 Q / 8 KV configuration as Qwen3-1.7B but is more heavily shaped by strong-to-weak distillation from larger teachers, which is explicitly designed to transfer both reasoning skills and mode-switching into lightweight models \cite{yang2025qwen3}. This combination of fine-grained attention, long-context training, and distillation from stronger "thinking/non-thinking" flagship models offers a plausible explanation for why Qwen3-4B reaches 95.64\% accuracy with RAG and why the 0.6B variant jumps to 88.12\% despite weak zero-shot and few-shot scores.

By contrast, the Gemma3 models seem to benefit from a different architectural bias. Gemma3-4B and Gemma3-1B also use a modern Transformer stack with GQA and QK-Norm, but their attention is organized into interleaved local and global blocks tailored for long-context throughput (e.g., 5 local layers followed by 1 global layer) \cite{team2025gemma}. In our RAG pipeline, each query log is followed by a small neighborhood of retrieved logs; Gemma’s local blocks can focus on within-log structure (e.g., message, unit, path), while its global blocks integrate across the query-plus-neighbors window. This architecture may be well aligned with “clustered” context layouts where relevant evidence is packed into short spans, which helps explain why Gemma3-1B and Gemma3-4B exhibit large gains under RAG (85.28\% and 81.84\%) despite relatively poor zero-shot baselines. Their lower Q-head counts (8/4 and 4/1) compared to Qwen3 limit expressivity somewhat, but the local–global pattern still lets them exploit retrieved examples effectively once those examples are explicitly surfaced by FAISS.

The Llama3.2 models, in turn, exemplify a more general-purpose small-model design. Public reports describe Llama 3–style models as decoder-only Transformers with GQA, long-context RoPE, and RMSNorm, trained on a large but more general mixture of web, code, and text than Qwen3’s STEM- and reasoning-heavy mix \cite{meta_llama3.2_docs_2024, grattafiori2024llama}. In our benchmarks, Llama3.2-3B and Llama3.2-1B have relatively high numbers of query heads (24 and 32, respectively) and 128K context, which supports efficient scanning of long inputs but without the same depth of reasoning-oriented post-training or strong-to-weak distillation. This pattern is consistent with their observed behavior: they benefit modestly from RAG (e.g., Llama3.2-3B rises to 53.31\% accuracy) but never reach the Qwen3/Gemma3 tier. Their architecture appears more suited to broad instruction following than to aggressively leveraging structured retrieved evidence in a narrow technical domain like syslog classification.

Finally, the degradation observed for several SRLMs under RAG, such as Qwen3-1.7B, DeepSeek-R1-Distill-Qwen-1.5B, and Phi-4-Mini-Reasoning may be interpreted through differences in architectural and post-training choices. Qwen3-1.7B shares the same 16 Q / 8 KV head configuration as Qwen3-0.6B, but receives heavier reasoning reinforcement learning (RL) and long-CoT training, which may bias it toward relying on its internally generated reasoning traces rather than on short, label-bearing retrieved snippets \cite{yang2025qwen3}. Similarly, DeepSeek-R1-Distill-Qwen-1.5B and Phi-4-Mini-Reasoning are distilled from larger reasoning models and explicitly optimized to produce long chain-of-thought outputs \cite{guo2025deepseek, xu2025phi}. Because our setup restricts the models to a single-digit output and supplies dense retrieval context, one plausible explanation is that CoT-centric models may over-weight internal reasoning, under-use retrieved examples, and sometimes emit verbose or malformed outputs, leading to degradation. This behavior is consistent with what we observe: high latency, frequent violation of the output format, and in some cases, sharp drops in accuracy when RAG is employed. Together, these patterns suggest that architectures and post-training pipelines geared toward controlled, retrieval-aware non-thinking modes may be better aligned with a RAG approach than the other configurations evaluated here.

\section{Conclusion and Future Work}\label{conclusionandfw}

This research treats severity classification as a controlled setting for probing retrieval behavior. The broader goal is to support diagnostic reasoning tasks such as RCA within DT environments, where compact models must integrate external evidence in real time. Our results show that SRLMs hold an advantage over SLMs in zero-shot and few-shot settings, while retrieval consistently strengthens SLM performance. However, retrieval benefits are not uniform: several SRLMs exhibit performance degradation under RAG, indicating that retrieval effectiveness depends not only on model size but also on architecture, training objectives, and the interaction between external context and internal reasoning dynamics.

To clarify the scope of these findings, this study does not claim that system log severity classification, as defined by Syslog labels, constitutes a comprehensive measure of log understanding or diagnostic competence. Severity levels are inherently noisy, administrator-defined, and weakly standardized across systems, which limits their suitability as ground-truth representations of underlying system state. Instead, severity classification is intentionally employed here as a constrained probing task that isolates a model’s ability to ground unstructured operational log content to semantically meaningful categories under strict output constraints. Accordingly, the reported accuracy and latency results should be interpreted as indicators of retrieval integration behavior and context utilization efficiency, rather than as direct proxies for correctness in downstream operational tasks such as RCA or anomaly diagnosis.

Within this landscape, Qwen3 provides a notable contrast to Llama3.2 and Gemma3. Qwen3 models are distilled from larger teacher models with an explicit thinking mode and trained on broad, multi-domain corpora, whereas Llama3.2 and Gemma3 prioritize instruction tuning and efficiency. Qwen3-4B, in particular, strikes a balance between capacity and stability: it has sufficient representational depth to parse noisy system logs and exploit retrieved top-$k$ neighbors without the instability observed in smaller variants. This is reflected in its substantial RAG improvement (56.01\% → 95.64\%), while Qwen3-1.7B degrades and Qwen3-0.6B relies heavily on near-label exemplars. These observations do not imply that SRLMs bias internal reasoning tokens, but they suggest that reasoning-oriented design and long-context training may shape how models integrate external evidence.

Interpreting these behaviors requires caution. Prior work on long-context limitations \cite{liu2024lost}, unfaithful or inconsistent chain-of-thought processes \cite{lanham2023measuring, turpin2023language}, and retrieval-aware training frameworks such as RA-DIT \cite{lin2023ra}, Self-RAG \cite{asai2024selfrag}, and IRCoT \cite{trivedi2023interleaving} offer plausible explanations and mechanisms for the mixed SRLM outcomes observed here. Our results should therefore be read as evidence of a potential tension between generative reasoning and retrieval, not a definitive claim about SRLM behavior. Future work will investigate whether these effects stem primarily from capacity constraints, training objectives, or deeper architectural dynamics. We further plan to extend this benchmark beyond system logs to other domain-specific tasks to test whether these retrieval patterns generalize across domains. Finally, integrating this framework into a live DT system will enable evaluation under streaming telemetry, shifting baselines, and evolving system state, where dynamic memory policies, time-aware retrieval, and continual context updates are essential for real-time diagnostic reasoning.

\paragraph{Author Contributions.} Conceptualization, C.Y. and Y.M.; methodology, Y.M., E.M., Z.W.; software, Y.M.; validation, Y.M. and Z.W.; resources, C.Y. and Z.W.; data curation, J.R., E.M., Y.M.; writing—original draft preparation, Y.M. and E.M.; writing—review and editing, Y.M., E.M., Z.W., C.Y.; supervision, C.Y.; project administration, C.Y.; funding acquisition, C.Y. All authors have read and agreed to the published version of the manuscript.

\paragraph{Funding.} This research was funded by NSF I/UCRC program (1841520); NASA Goddard CISTO; and the ASSIP program, supported by George Mason University’s College of Science.

\paragraph{Data Availability Statement.} The original data and source code presented in this study are openly available at https://github.com/stccenter/Benchmarking-SLMs-and-SRLMs-on-System-Log-Severity-Classification

\paragraph{Acknowledgments.} We are grateful to Anusha Srirenganathan Malarvizhi and Yaya Lan for helpful discussions and advice.

\paragraph{Conflicts of Interest.} The authors declare no conflict of interest.

\bibliographystyle{unsrtnat}
\bibliography{refs}

\end{document}

%% file: macros.tex
\usepackage{fullpage} % small margins
\usepackage{microtype}
\usepackage{graphicx}
\usepackage{subfigure}
\usepackage{booktabs} 
\usepackage{color}
\usepackage{lmodern}

\usepackage{natbib}
\usepackage{hyperref}

\usepackage{amsmath}
\usepackage{amssymb}  
\usepackage{mathtools}
\usepackage{amsthm}
 \usepackage[capitalize]{cleveref}
\usepackage{bm}
\usepackage{listings}

\crefname{observation}{Observation}{Observations}
\Crefname{equation}{Eq.}{Eqs.}
\Crefname{figure}{Fig.}{Figs.}
\Crefname{table}{Table}{Tables}

\DeclareMathOperator*{\argmin}{arg\,min}

\def\1{\mathbf{1}}

\newcommand{\prompt}[1]{\verb+#1+}

%% file: refs.bib
@article{gholamian2021comprehensive,
  title={A comprehensive survey of logging in software: From logging statements automation to log mining and analysis},
  author={Gholamian, Sina and Ward, Paul AS},
  journal={arXiv preprint arXiv:2110.12489},
  year={2021}
}

@inproceedings{du2017deeplog,
  title={Deeplog: Anomaly detection and diagnosis from system logs through deep learning},
  author={Du, Min and Li, Feifei and Zheng, Guineng and Srikumar, Vivek},
  booktitle={Proceedings of the 2017 ACM SIGSAC conference on computer and communications security},
  pages={1285--1298},
  year={2017}
}

@inproceedings{sekar2024eaudit,
  title={eaudit: A fast, scalable and deployable audit data collection system},
  author={Sekar, R and Kimm, Hanke and Aich, Rohit},
  booktitle={2024 IEEE Symposium on Security and Privacy (SP)},
  pages={3571--3589},
  year={2024},
  organization={IEEE}
}

@article{albert2025system,
  title={System Logs Anomaly Detection. Are we on the right path?},
  author={Albert, Ramona-Georgiana},
  journal={Applied Artificial Intelligence},
  volume={39},
  number={1},
  pages={2440692},
  year={2025},
  publisher={Taylor \& Francis}
}

@article{viola2022combining,
  title={Combining log files and monitoring data to detect anomaly patterns in a data center},
  author={Viola, Laura and Ronchieri, Elisabetta and Cavallaro, Claudia},
  journal={Computers},
  volume={11},
  number={8},
  pages={117},
  year={2022},
  publisher={MDPI}
}

@inproceedings{dietz2020integrating,
  title={Integrating digital twin security simulations in the security operations center},
  author={Dietz, Marietheres and Vielberth, Manfred and Pernul, G{\"u}nther},
  booktitle={Proceedings of the 15th International Conference on Availability, Reliability and Security},
  pages={1--9},
  year={2020}
}

@inproceedings{zhang2017syslog,
  title={Syslog processing for switch failure diagnosis and prediction in datacenter networks},
  author={Zhang, Shenglin and Meng, Weibin and Bu, Jiahao and Yang, Sen and Liu, Ying and Pei, Dan and Xu, Jun and Chen, Yu and Dong, Hui and Qu, Xianping and others},
  booktitle={2017 IEEE/ACM 25th International Symposium on Quality of Service (IWQoS)},
  pages={1--10},
  year={2017},
  organization={IEEE}
}

@inproceedings{xu2009detecting,
  title={Detecting large-scale system problems by mining console logs},
  author={Xu, Wei and Huang, Ling and Fox, Armando and Patterson, David and Jordan, Michael I},
  booktitle={Proceedings of the ACM SIGOPS 22nd symposium on Operating systems principles},
  pages={117--132},
  year={2009}
}

@inproceedings{jiang2009understanding,
  title={Understanding Customer Problem Troubleshooting from Storage System Logs.},
  author={Jiang, Weihang and Hu, Chongfeng and Pasupathy, Shankar and Kanevsky, Arkady and Li, Zhenmin and Zhou, Yuanyuan},
  booktitle={FAST},
  volume={9},
  pages={43--56},
  year={2009}
}

@article{landauer2023deep,
  title={Deep learning for anomaly detection in log data: A survey},
  author={Landauer, Max and Onder, Sebastian and Skopik, Florian and Wurzenberger, Markus},
  journal={Machine Learning with Applications},
  volume={12},
  pages={100470},
  year={2023},
  publisher={Elsevier}
}

@inproceedings{gupta2023learning,
  title={Learning representations on logs for aiops},
  author={Gupta, Pranjal and Kumar, Harshit and Kar, Debanjana and Bhukar, Karan and Aggarwal, Pooja and Mohapatra, Prateeti},
  booktitle={2023 IEEE 16th International Conference on Cloud Computing (CLOUD)},
  pages={155--166},
  year={2023},
  organization={IEEE}
}

@inproceedings{mahindru2021log,
  title={Log anomaly to resolution: Ai based proactive incident remediation},
  author={Mahindru, Ruchi and Kumar, Harshit and Bansal, Sahil},
  booktitle={2021 36th IEEE/ACM International Conference on Automated Software Engineering (ASE)},
  pages={1353--1357},
  year={2021},
  organization={IEEE}
}

@inproceedings{bansal2020decaf,
  title={Decaf: Diagnosing and triaging performance issues in large-scale cloud services},
  author={Bansal, Chetan and Renganathan, Sundararajan and Asudani, Ashima and Midy, Olivier and Janakiraman, Mathru},
  booktitle={Proceedings of the ACM/IEEE 42nd International Conference on Software Engineering: Software Engineering in Practice},
  pages={201--210},
  year={2020}
}

@article{sun2025accurate,
  title={Accurate and Interpretable Log-Based Fault Diagnosis using Large Language Models},
  author={Sun, Yongqian and Ma, Shiyu and Xiao, Tong and Zhao, Yongxin and Cai, Xuhui and Dong, Wei and Shen, Yue and Zhao, Yao and Zhang, Shenglin and Han, Jing and others},
  journal={IEEE Transactions on Services Computing},
  year={2025},
  publisher={IEEE}
}

@article{wang2025optimizing,
  title={Optimizing context-based location extraction by tuning open-source LLMs with RAG},
  author={Wang, Zifu and Masri, Yahya and Malarvizhi, Anusha Srirenganathan and Stover, Tayven and Ahmed, Samir and Wong, David and Jiang, Yongyao and Li, Yun and Bere, Mathieu and Rothbart, Daniel and others},
  journal={International Journal of Digital Earth},
  volume={18},
  number={1},
  pages={2521786},
  year={2025},
  publisher={Taylor \& Francis}
}

@article{masri2025comparative,
  title={Comparative Analysis of BERT and GPT for Classifying Crisis News with Sudan Conflict as an Example},
  author={Masri, Yahya and Wang, Zifu and Srirenganathan Malarvizhi, Anusha and Ahmed, Samir and Stover, Tayven and Wong, David WS and Jiang, Yongyao and Li, Yun and Liu, Qian and Bere, Mathieu and others},
  journal={Algorithms},
  volume={18},
  number={7},
  pages={420},
  year={2025},
  publisher={MDPI}
}

@article{setty2024improving,
  title={Improving retrieval for rag based question answering models on financial documents},
  author={Setty, Spurthi and Thakkar, Harsh and Lee, Alyssa and Chung, Eden and Vidra, Natan},
  journal={arXiv preprint arXiv:2404.07221},
  year={2024}
}

@inproceedings{yang2021semi,
  title={Semi-supervised log-based anomaly detection via probabilistic label estimation},
  author={Yang, Lin and Chen, Junjie and Wang, Zan and Wang, Weijing and Jiang, Jiajun and Dong, Xuyuan and Zhang, Wenbin},
  booktitle={2021 IEEE/ACM 43rd International Conference on Software Engineering (ICSE)},
  pages={1448--1460},
  year={2021},
  organization={IEEE}
}

@article{chen2022bert,
  title={Bert-log: Anomaly detection for system logs based on pre-trained language model},
  author={Chen, Song and Liao, Hai},
  journal={Applied Artificial Intelligence},
  volume={36},
  number={1},
  pages={2145642},
  year={2022},
  publisher={Taylor \& Francis}
}

@article{landauer2020system,
  title={System log clustering approaches for cyber security applications: A survey},
  author={Landauer, Max and Skopik, Florian and Wurzenberger, Markus and Rauber, Andreas},
  journal={Computers \& Security},
  volume={92},
  pages={101739},
  year={2020},
  publisher={Elsevier}
}

@misc{gerhards2009rfc,
  title={RFC 5424: The syslog protocol},
  author={Gerhards, Rainer},
  year={2009},
  publisher={RFC Editor}
}

@misc{lonvick2001rfc3164,
  title={Rfc3164: The bsd syslog protocol},
  author={Lonvick, Chris},
  year={2001},
  publisher={RFC Editor}
}

@article{el2020systematic,
  title={A systematic literature review on automated log abstraction techniques},
  author={El-Masri, Diana and Petrillo, Fabio and Gu{\'e}h{\'e}neuc, Yann-Ga{\"e}l and Hamou-Lhadj, Abdelwahab and Bouziane, Anas},
  journal={Information and Software Technology},
  volume={122},
  pages={106276},
  year={2020},
  publisher={Elsevier}
}

@inproceedings{li2021deeplv,
  title={Deeplv: Suggesting log levels using ordinal based neural networks},
  author={Li, Zhenhao and Li, Heng and Chen, Tse-Hsun and Shang, Weiyi},
  booktitle={2021 IEEE/ACM 43rd International Conference on Software Engineering (ICSE)},
  pages={1461--1472},
  year={2021},
  organization={IEEE}
}

@inproceedings{zhu2019tools,
  title={Tools and benchmarks for automated log parsing},
  author={Zhu, Jieming and He, Shilin and Liu, Jinyang and He, Pinjia and Xie, Qi and Zheng, Zibin and Lyu, Michael R},
  booktitle={2019 IEEE/ACM 41st International Conference on Software Engineering: Software Engineering in Practice (ICSE-SEIP)},
  pages={121--130},
  year={2019},
  organization={Ieee}
}

@article{meng2021logclass,
  title={Logclass: Anomalous log identification and classification with partial labels},
  author={Meng, Weibin and Liu, Ying and Zhang, Shenglin and Zaiter, Federico and Zhang, Yuzhe and Huang, Yuheng and Yu, Zhaoyang and Zhang, Yuzhi and Song, Lei and Zhang, Ming and others},
  journal={IEEE Transactions on Network and Service Management},
  volume={18},
  number={2},
  pages={1870--1884},
  year={2021},
  publisher={IEEE}
}

@inproceedings{he2016experience,
  title={Experience report: System log analysis for anomaly detection},
  author={He, Shilin and Zhu, Jieming and He, Pinjia and Lyu, Michael R},
  booktitle={2016 IEEE 27th international symposium on software reliability engineering (ISSRE)},
  pages={207--218},
  year={2016},
  organization={IEEE}
}

@inproceedings{dixit2022utilizing,
  title={Utilizing ML and DL algorithms for alert classification in intrusion detection and prevention systems: A detailed review},
  author={Dixit, Utkarsh and Bhatia, Suman and Bhatia, Pramod},
  booktitle={2022 2nd International Conference on Advance Computing and Innovative Technologies in Engineering (ICACITE)},
  pages={1199--1205},
  year={2022},
  organization={IEEE}
}

@article{ali2025comprehensive,
  title={A comprehensive study of machine learning techniques for log-based anomaly detection},
  author={Ali, Shan and Boufaied, Chaima and Bianculli, Domenico and Branco, Paula and Briand, Lionel},
  journal={Empirical Software Engineering},
  volume={30},
  number={5},
  pages={129},
  year={2025},
  publisher={Springer}
}

@inproceedings{qi2022adanomaly,
  title={Adanomaly: adaptive anomaly detection for system logs with adversarial learning},
  author={Qi, Jiaxing and Luan, Zhongzhi and Huang, Shaohan and Wang, Yukun and Fung, Carol and Yang, Hailong and Qian, Depei},
  booktitle={NOMS 2022-2022 IEEE/IFIP Network Operations and Management Symposium},
  pages={1--5},
  year={2022},
  organization={IEEE}
}

@article{li2022swisslog,
  title={SwissLog: Robust anomaly detection and localization for interleaved unstructured logs},
  author={Li, Xiaoyun and Chen, Pengfei and Jing, Linxiao and He, Zilong and Yu, Guangba},
  journal={IEEE Transactions on Dependable and Secure Computing},
  volume={20},
  number={4},
  pages={2762--2780},
  year={2022},
  publisher={IEEE}
}

@article{chen2021experience,
  title={Experience report: Deep learning-based system log analysis for anomaly detection},
  author={Chen, Zhuangbin and Liu, Jinyang and Gu, Wenwei and Su, Yuxin and Lyu, Michael R},
  journal={arXiv preprint arXiv:2107.05908},
  year={2021}
}

@article{liu2023logbd,
  title={LogBD: A log anomaly detection method based on pretrained models and domain adaptation},
  author={Liu, Shuxian and Deng, Le and Xu, Huan and Wang, Wei},
  journal={Applied Sciences},
  volume={13},
  number={13},
  pages={7739},
  year={2023},
  publisher={MDPI}
}

@article{duan2024logedl,
  title={LogEDL: Log Anomaly Detection via Evidential Deep Learning},
  author={Duan, Yunfeng and Xue, Kaiwen and Sun, Hao and Bao, Haotong and Wei, Yadong and You, Zhangzheng and Zhang, Yuantian and Jiang, Xiwei and Yang, Sangning and Chen, Jiaxing and others},
  journal={Applied Sciences},
  volume={14},
  number={16},
  pages={7055},
  year={2024},
  publisher={MDPI}
}

@inproceedings{guo2024logformer,
  title={Logformer: A pre-train and tuning pipeline for log anomaly detection},
  author={Guo, Hongcheng and Yang, Jian and Liu, Jiaheng and Bai, Jiaqi and Wang, Boyang and Li, Zhoujun and Zheng, Tieqiao and Zhang, Bo and Peng, Junran and Tian, Qi},
  booktitle={Proceedings of the AAAI conference on artificial intelligence},
  volume={38},
  pages={135--143},
  year={2024}
}

@article{alzu2025cyberattack,
  title={Cyberattack event logs classification using deep learning with semantic feature analysis},
  author={Alzu’bi, Ahmad and Darwish, Omar and Albashayreh, Amjad and Tashtoush, Yahya},
  journal={Computers \& Security},
  volume={150},
  pages={104222},
  year={2025},
  publisher={Elsevier}
}

@inproceedings{yuan2020ada,
  title={Ada: Adaptive deep log anomaly detector},
  author={Yuan, Yali and Adhatarao, Sripriya Srikant and Lin, Mingkai and Yuan, Yachao and Liu, Zheli and Fu, Xiaoming},
  booktitle={Ieee Infocom 2020-ieee Conference on Computer Communications},
  pages={2449--2458},
  year={2020},
  organization={IEEE}
}

@article{henriques2020combining,
  title={Combining k-means and xgboost models for anomaly detection using log datasets},
  author={Henriques, Jo{\~a}o and Caldeira, Filipe and Cruz, Tiago and Sim{\~o}es, Paulo},
  journal={Electronics},
  volume={9},
  number={7},
  pages={1164},
  year={2020},
  publisher={MDPI}
}

@inproceedings{zhang2024end,
  title={End-to-end automl for unsupervised log anomaly detection},
  author={Zhang, Shenglin and Ji, Yuhe and Luan, Jiaqi and Nie, Xiaohui and Chen, Ziang and Ma, Minghua and Sun, Yongqian and Pei, Dan},
  booktitle={Proceedings of the 39th IEEE/ACM International Conference on Automated Software Engineering},
  pages={1680--1692},
  year={2024}
}

@inproceedings{li2024graph,
  title={Graph neural networks based log anomaly detection and explanation},
  author={Li, Zhong and Shi, Jiayang and Van Leeuwen, Matthijs},
  booktitle={Proceedings of the 2024 IEEE/ACM 46th international conference on software engineering: companion proceedings},
  pages={306--307},
  year={2024}
}

@inproceedings{wang2021multi,
  title={Multi-scale one-class recurrent neural networks for discrete event sequence anomaly detection},
  author={Wang, Zhiwei and Chen, Zhengzhang and Ni, Jingchao and Liu, Hui and Chen, Haifeng and Tang, Jiliang},
  booktitle={Proceedings of the 27th ACM SIGKDD conference on knowledge discovery \& data mining},
  pages={3726--3734},
  year={2021}
}

@inproceedings{yu2024deep,
  title={Deep learning or classical machine learning? an empirical study on log-based anomaly detection},
  author={Yu, Boxi and Yao, Jiayi and Fu, Qiuai and Zhong, Zhiqing and Xie, Haotian and Wu, Yaoliang and Ma, Yuchi and He, Pinjia},
  booktitle={Proceedings of the 46th IEEE/ACM international conference on software engineering},
  pages={1--13},
  year={2024}
}

@article{zhang2023system,
  title={System log parsing: A survey},
  author={Zhang, Tianzhu and Qiu, Han and Castellano, Gabriele and Rifai, Myriana and Chen, Chung Shue and Pianese, Fabio},
  journal={IEEE Transactions on Knowledge and Data Engineering},
  volume={35},
  number={8},
  pages={8596--8614},
  year={2023},
  publisher={IEEE}
}

@article{bhanage2021infrastructure,
  title={IT infrastructure anomaly detection and failure handling: A systematic literature review focusing on datasets, log preprocessing, machine \& deep learning approaches and automated tool},
  author={Bhanage, Deepali Arun and Pawar, Ambika Vishal and Kotecha, Ketan},
  journal={IEEE Access},
  volume={9},
  pages={156392--156421},
  year={2021},
  publisher={IEEE}
}

@article{he2021survey,
  title={A survey on automated log analysis for reliability engineering},
  author={He, Shilin and He, Pinjia and Chen, Zhuangbin and Yang, Tianyi and Su, Yuxin and Lyu, Michael R},
  journal={ACM computing surveys (CSUR)},
  volume={54},
  number={6},
  pages={1--37},
  year={2021},
  publisher={ACM New York, NY, USA}
}

@inproceedings{le2022log,
  title={Log-based anomaly detection with deep learning: How far are we?},
  author={Le, Van-Hoang and Zhang, Hongyu},
  booktitle={Proceedings of the 44th international conference on software engineering},
  pages={1356--1367},
  year={2022}
}

@article{yang2025logllama,
  title={LogLLaMA: Transformer-based log anomaly detection with LLaMA},
  author={Yang, Zhuoyi and Harris, Ian G},
  journal={arXiv preprint arXiv:2503.14849},
  year={2025}
}

@article{zhang2025novel,
  title={A Novel GPT-Based Framework for Anomaly Detection in System Logs},
  author={Zhang, Zeng and Yin, Wenjie and Li, Xiaoqi},
  journal={arXiv preprint arXiv:2510.16044},
  year={2025}
}

@inproceedings{huang2025logrules,
  title={LogRules: Enhancing Log Analysis Capability of Large Language Models through Rules},
  author={Huang, Xin and Zhang, Ting and Zhao, Wen},
  booktitle={Findings of the Association for Computational Linguistics: NAACL 2025},
  pages={452--470},
  year={2025}
}

@inproceedings{heng2025benchmarking,
  title={Benchmarking Open-Source Large Language Models for Log Level Suggestion},
  author={Heng, Yi Wen and Ma, Zeyang and Li, Zhenhao and Kim, Dong Jae and Chen, Tse-Hsun},
  booktitle={2025 IEEE Conference on Software Testing, Verification and Validation (ICST)},
  pages={314--325},
  year={2025},
  organization={IEEE}
}

@article{ouatiti2025omnillp,
  title={OmniLLP: Enhancing LLM-based Log Level Prediction with Context-Aware Retrieval},
  author={Ouatiti, Youssef Esseddiq and Sayagh, Mohammed and Adams, Bram and Hassan, Ahmed E},
  journal={arXiv preprint arXiv:2508.08545},
  year={2025}
}

@article{karlsen2024benchmarking,
  title={Benchmarking large language models for log analysis, security, and interpretation},
  author={Karlsen, Egil and Luo, Xiao and Zincir-Heywood, Nur and Heywood, Malcolm},
  journal={Journal of Network and Systems Management},
  volume={32},
  number={3},
  pages={59},
  year={2024},
  publisher={Springer}
}

@article{ihalage2025convolutional,
  title={Convolutional vs large language models for software log classification in edge-deployable cellular network testing},
  author={Ihalage, Achintha and Taheri, Sayed and Muhammad, Faris and Al-Raweshidy, Hamed},
  journal={IEEE Access},
  year={2025},
  publisher={IEEE}
}

@inproceedings{liu2024interpretable,
  title={Interpretable online log analysis using large language models with prompt strategies},
  author={Liu, Yilun and Tao, Shimin and Meng, Weibin and Wang, Jingyu and Ma, Wenbing and Chen, Yuhang and Zhao, Yanqing and Yang, Hao and Jiang, Yanfei},
  booktitle={Proceedings of the 32nd IEEE/ACM International Conference on Program Comprehension},
  pages={35--46},
  year={2024}
}

@article{ocansey2025logtinyllm,
  title={LogTinyLLM: Tiny Large Language Models Based Contextual Log Anomaly Detection},
  author={Ocansey, Isaiah Thompson and Bhattacharya, Ritwik and Sen, Tanmay},
  journal={arXiv preprint arXiv:2507.11071},
  year={2025}
}

@article{kim2025automated,
  title={Automated Log Statement using Source-Code Metrics},
  author={Kim, Se-Jin and Lee, Chan-Gun},
  journal={IEEE Access},
  year={2025},
  publisher={IEEE}
}

@article{yang2025large,
  title={Large language models for network intrusion detection systems: Foundations, implementations, and future directions},
  author={Yang, Shuo and Zheng, Xinran and Zhang, Xinchen and Xu, Jinfeng and Li, Jinze and Xie, Donglin and Long, Weicai and Ngai, Edith CH},
  journal={arXiv preprint arXiv:2507.04752},
  year={2025}
}

@article{bui2024systematic,
  title={A Systematic Comparison of Large Language Models Performance for Intrusion Detection},
  author={Bui, Minh-Thanh and Boffa, Matteo and Valentim, Rodolfo Vieira and Navarro, Jose Manuel and Chen, Fuxing and Bao, Xiaosheng and Houidi, Zied Ben and Rossi, Dario},
  journal={Proceedings of the ACM on Networking},
  volume={2},
  number={CoNEXT4},
  pages={1--23},
  year={2024},
  publisher={ACM New York, NY, USA}
}

@misc{ahmed2025optimizing,
  title={Optimizing LLMs for Microservices Logs Analysis through Prompt Engineering},
  author={Ahmed, Sahar},
  year={2025}
}

@inproceedings{liu2024logprompt,
  title={Logprompt: Prompt engineering towards zero-shot and interpretable log analysis},
  author={Liu, Yilun and Tao, Shimin and Meng, Weibin and Yao, Feiyu and Zhao, Xiaofeng and Yang, Hao},
  booktitle={Proceedings of the 2024 IEEE/ACM 46th International Conference on Software Engineering: Companion Proceedings},
  pages={364--365},
  year={2024}
}

@article{cui2024logeval,
  title={Logeval: A comprehensive benchmark suite for large language models in log analysis},
  author={Cui, Tianyu and Ma, Shiyu and Chen, Ziang and Xiao, Tong and Tao, Shimin and Liu, Yilun and Zhang, Shenglin and Lin, Duoming and Liu, Changchang and Cai, Yuzhe and others},
  journal={arXiv preprint arXiv:2407.01896},
  year={2024}
}

@inproceedings{duan2025eagerlog,
  title={EagerLog: Active Learning Enhanced Retrieval Augmented Generation for Log-based Anomaly Detection},
  author={Duan, Chiming and Jia, Tong and Yang, Yong and Liu, Guiyang and Liu, Jinbu and Zhang, Huxing and Zhou, Qi and Li, Ying and Huang, Gang},
  booktitle={ICASSP 2025-2025 IEEE International Conference on Acoustics, Speech and Signal Processing (ICASSP)},
  pages={1--5},
  year={2025},
  organization={IEEE}
}

@article{sharma2025retrieval,
  title={Retrieval-Augmented Generation: A Comprehensive Survey of Architectures, Enhancements, and Robustness Frontiers},
  author={Sharma, Chaitanya},
  journal={arXiv preprint arXiv:2506.00054},
  year={2025}
}

@inproceedings{wu2025multirag,
  title={Multirag: a knowledge-guided framework for mitigating hallucination in multi-source retrieval augmented generation},
  author={Wu, Wenlong and Wang, Haofen and Li, Bohan and Huang, Peixuan and Zhao, Xinzhe and Liang, Lei},
  booktitle={2025 IEEE 41st International Conference on Data Engineering (ICDE)},
  pages={3070--3083},
  year={2025},
  organization={IEEE}
}

@article{yang2025qwen3,
  title={Qwen3 technical report},
  author={Yang, An and Li, Anfeng and Yang, Baosong and Zhang, Beichen and Hui, Binyuan and Zheng, Bo and Yu, Bowen and Gao, Chang and Huang, Chengen and Lv, Chenxu and others},
  journal={arXiv preprint arXiv:2505.09388},
  year={2025}
}

@article{team2025gemma,
  title={Gemma 3 technical report},
  author={Team, Gemma and Kamath, Aishwarya and Ferret, Johan and Pathak, Shreya and Vieillard, Nino and Merhej, Ramona and Perrin, Sarah and Matejovicova, Tatiana and Ram{\'e}, Alexandre and Rivi{\`e}re, Morgane and others},
  journal={arXiv preprint arXiv:2503.19786},
  year={2025}
}

@misc{meta_llama3.2_docs_2024,
  title        = {Llama 3.2: Model Cards and Prompt Formats},
  author       = {Meta AI},
  year         = {2024},
  howpublished = {\url{https://www.llama.com/docs/model-cards-and-prompt-formats/llama3_2/}},
  note         = {Accessed: 2025-12-04}
}

@article{grattafiori2024llama,
  title={The llama 3 herd of models},
  author={Grattafiori, Aaron and Dubey, Abhimanyu and Jauhri, Abhinav and Pandey, Abhinav and Kadian, Abhishek and Al-Dahle, Ahmad and Letman, Aiesha and Mathur, Akhil and Schelten, Alan and Vaughan, Alex and others},
  journal={arXiv preprint arXiv:2407.21783},
  year={2024}
}

@article{guo2025deepseek,
  title={Deepseek-r1: Incentivizing reasoning capability in llms via reinforcement learning},
  author={Guo, Daya and Yang, Dejian and Zhang, Haowei and Song, Junxiao and Zhang, Ruoyu and Xu, Runxin and Zhu, Qihao and Ma, Shirong and Wang, Peiyi and Bi, Xiao and others},
  journal={arXiv preprint arXiv:2501.12948},
  year={2025}
}

@article{xu2025phi,
  title={Phi-4-mini-reasoning: Exploring the limits of small reasoning language models in math},
  author={Xu, Haoran and Peng, Baolin and Awadalla, Hany and Chen, Dongdong and Chen, Yen-Chun and Gao, Mei and Kim, Young Jin and Li, Yunsheng and Ren, Liliang and Shen, Yelong and others},
  journal={arXiv preprint arXiv:2504.21233},
  year={2025}
}

@inproceedings{asai2024selfrag,
  title        = {SELF-RAG: Learning to Retrieve, Generate, and Critique through Self-Reflection},
  author       = {Asai, Akari and Wu, Zeqiu and Wang, Yizhong and Sil, Avirup and Hajishirzi, Hannaneh},
  booktitle    = {Proceedings of the International Conference on Learning Representations (ICLR)},
  year         = {2024},
  url          = {https://selfrag.github.io/}
}

@inproceedings{lin2023ra,
  title={Ra-dit: Retrieval-augmented dual instruction tuning},
  author={Lin, Xi Victoria and Chen, Xilun and Chen, Mingda and Shi, Weijia and Lomeli, Maria and James, Richard and Rodriguez, Pedro and Kahn, Jacob and Szilvasy, Gergely and Lewis, Mike and others},
  booktitle={The Twelfth International Conference on Learning Representations},
  year={2023}
}

@article{lanham2023measuring,
  title={Measuring faithfulness in chain-of-thought reasoning},
  author={Lanham, Tamera and Chen, Anna and Radhakrishnan, Ansh and Steiner, Benoit and Denison, Carson and Hernandez, Danny and Li, Dustin and Durmus, Esin and Hubinger, Evan and Kernion, Jackson and others},
  journal={arXiv preprint arXiv:2307.13702},
  year={2023}
}

@article{liu2024lost,
  title={Lost in the middle: How language models use long contexts},
  author={Liu, Nelson F and Lin, Kevin and Hewitt, John and Paranjape, Ashwin and Bevilacqua, Michele and Petroni, Fabio and Liang, Percy},
  journal={Transactions of the Association for Computational Linguistics},
  volume={12},
  pages={157--173},
  year={2024}
}

@inproceedings{trivedi2023interleaving,
  title={Interleaving retrieval with chain-of-thought reasoning for knowledge-intensive multi-step questions},
  author={Trivedi, Harsh and Balasubramanian, Niranjan and Khot, Tushar and Sabharwal, Ashish},
  booktitle={Proceedings of the 61st annual meeting of the association for computational linguistics (volume 1: long papers)},
  pages={10014--10037},
  year={2023}
}

@article{turpin2023language,
  title={Language models don't always say what they think: Unfaithful explanations in chain-of-thought prompting},
  author={Turpin, Miles and Michael, Julian and Perez, Ethan and Bowman, Samuel},
  journal={Advances in Neural Information Processing Systems},
  volume={36},
  pages={74952--74965},
  year={2023}
}

@article{braun2025hidden,
  title={The Hidden Structure--Improving Legal Document Understanding Through Explicit Text Formatting},
  author={Braun, Christian and Lilienbeck, Alexander and Mentjukov, Daniel},
  journal={arXiv preprint arXiv:2505.12837},
  year={2025}
}

@article{he2024does,
  title={Does prompt formatting have any impact on llm performance?},
  author={He, Jia and Rungta, Mukund and Koleczek, David and Sekhon, Arshdeep and Wang, Franklin X and Hasan, Sadid},
  journal={arXiv preprint arXiv:2411.10541},
  year={2024}
}

@inproceedings{kong2024better,
  title={Better zero-shot reasoning with role-play prompting},
  author={Kong, Aobo and Zhao, Shiwan and Chen, Hao and Li, Qicheng and Qin, Yong and Sun, Ruiqi and Zhou, Xin and Wang, Enzhi and Dong, Xiaohang},
  booktitle={Proceedings of the 2024 Conference of the North American Chapter of the Association for Computational Linguistics: Human Language Technologies (Volume 1: Long Papers)},
  pages={4099--4113},
  year={2024}
}

@inproceedings{dong2024survey,
  title={A survey on in-context learning},
  author={Dong, Qingxiu and Li, Lei and Dai, Damai and Zheng, Ce and Ma, Jingyuan and Li, Rui and Xia, Heming and Xu, Jingjing and Wu, Zhiyong and Chang, Baobao and others},
  booktitle={Proceedings of the 2024 conference on empirical methods in natural language processing},
  pages={1107--1128},
  year={2024}
}

@misc{lmstudio2025,
  author       = {{LM Studio Inc.}},
  title        = {LM Studio [Computer software]},
  year         = {2025},
  howpublished = {\url{https://lmstudio.ai}},
  note         = {Accessed: 29 October 2025}
}

@article{wang2025short,
  title={A Short Survey on Small Reasoning Models: Training, Inference, Applications and Research Directions},
  author={Wang, Chengyu and Zhang, Taolin and Hong, Richang and Huang, Jun},
  journal={arXiv preprint arXiv:2504.09100},
  year={2025}
}

@article{belcak2025small,
  title={Small Language Models are the Future of Agentic AI},
  author={Belcak, Peter and Heinrich, Greg and Diao, Shizhe and Fu, Yonggan and Dong, Xin and Muralidharan, Saurav and Lin, Yingyan Celine and Molchanov, Pavlo},
  journal={arXiv preprint arXiv:2506.02153},
  year={2025}
}

@inproceedings{wang2025survey,
  title={A survey on small language models in the era of large language models: Architecture, capabilities, and trustworthiness},
  author={Wang, Fali and Lin, Minhua and Ma, Yao and Liu, Hui and He, Qi and Tang, Xianfeng and Tang, Jiliang and Pei, Jian and Wang, Suhang},
  booktitle={Proceedings of the 31st ACM SIGKDD Conference on Knowledge Discovery and Data Mining V. 2},
  pages={6173--6183},
  year={2025}
}

@article{pham2025slm,
  title={SLM-Bench: A Comprehensive Benchmark of Small Language Models on Environmental Impacts--Extended Version},
  author={Pham, Nghiem Thanh and Kieu, Tung and Nguyen, Duc-Manh and Xuan, Son Ha and Duong-Trung, Nghia and Le-Phuoc, Danh},
  journal={arXiv preprint arXiv:2508.15478},
  year={2025}
}

@inproceedings{chitty2024llm,
  title={Llm-inference-bench: Inference benchmarking of large language models on ai accelerators},
  author={Chitty-Venkata, Krishna Teja and Raskar, Siddhisanket and Kale, Bharat and Ferdaus, Farah and Tanikanti, Aditya and Raffenetti, Ken and Taylor, Valerie and Emani, Murali and Vishwanath, Venkatram},
  booktitle={SC24-W: Workshops of the International Conference for High Performance Computing, Networking, Storage and Analysis},
  pages={1362--1379},
  year={2024},
  organization={IEEE}
}

@article{wang2023adopting,
  title={Adopting GPU computing to support DL-based Earth science applications},
  author={Wang, Zifu and Li, Yun and Wang, Kevin and Cain, Jacob and Salami, Mary and Duffy, Daniel Q and Little, Michael M and Yang, Chaowei},
  journal={International Journal of Digital Earth},
  volume={16},
  number={1},
  pages={2660--2680},
  year={2023},
  publisher={Taylor \& Francis}
}

@article{ma2025adaptivelog,
  title={Adaptivelog: An adaptive log analysis framework with the collaboration of large and small language model},
  author={Ma, Lipeng and Yang, Weidong and Li, Yixuan and Fei, Ben and Zhou, Mingjie and Li, Shuhao and Jiang, Sihang and Xu, Bo and Xiao, Yanghua},
  journal={ACM Transactions on Software Engineering and Methodology},
  year={2025},
  publisher={ACM New York, NY}
}

@article{radford2019language,
  title={Language models are unsupervised multitask learners},
  author={Radford, Alec and Wu, Jeffrey and Child, Rewon and Luan, David and Amodei, Dario and Sutskever, Ilya and others},
  journal={OpenAI blog},
  volume={1},
  number={8},
  pages={9},
  year={2019}
}

@article{kojima2022large,
  title={Large language models are zero-shot reasoners},
  author={Kojima, Takeshi and Gu, Shixiang Shane and Reid, Machel and Matsuo, Yutaka and Iwasawa, Yusuke},
  journal={Advances in neural information processing systems},
  volume={35},
  pages={22199--22213},
  year={2022}
}

@article{brown2020language,
  title={Language models are few-shot learners},
  author={Brown, Tom and Mann, Benjamin and Ryder, Nick and Subbiah, Melanie and Kaplan, Jared D and Dhariwal, Prafulla and Neelakantan, Arvind and Shyam, Pranav and Sastry, Girish and Askell, Amanda and others},
  journal={Advances in neural information processing systems},
  volume={33},
  pages={1877--1901},
  year={2020}
}

@article{lewis2020retrieval,
  title={Retrieval-augmented generation for knowledge-intensive nlp tasks},
  author={Lewis, Patrick and Perez, Ethan and Piktus, Aleksandra and Petroni, Fabio and Karpukhin, Vladimir and Goyal, Naman and K{\"u}ttler, Heinrich and Lewis, Mike and Yih, Wen-tau and Rockt{\"a}schel, Tim and others},
  journal={Advances in neural information processing systems},
  volume={33},
  pages={9459--9474},
  year={2020}
}

@article{douze2025faiss,
  title={The faiss library},
  author={Douze, Matthijs and Guzhva, Alexandr and Deng, Chengqi and Johnson, Jeff and Szilvasy, Gergely and Mazar{\'e}, Pierre-Emmanuel and Lomeli, Maria and Hosseini, Lucas and J{\'e}gou, Herv{\'e}},
  journal={IEEE Transactions on Big Data},
  year={2025},
  publisher={IEEE}
}

@article{ramachandran2023automated,
  title={Automated log classification using deep learning},
  author={Ramachandran, Shekar and Agrahari, Rupali and Mudgal, Priyanka and Bhilwaria, Harshita and Long, Garth and Kumar, Arisha},
  journal={Procedia Computer Science},
  volume={218},
  pages={1722--1732},
  year={2023},
  publisher={Elsevier}
}

@article{nussbaum2024nomic,
  title={Nomic embed: Training a reproducible long context text embedder},
  author={Nussbaum, Zach and Morris, John X and Duderstadt, Brandon and Mulyar, Andriy},
  journal={arXiv preprint arXiv:2402.01613},
  year={2024}
}
